\relax
\documentclass[letterpaper]{article} 
\usepackage{aaai22}  
\usepackage{times}  
\usepackage{helvet}  
\usepackage{courier}  
\usepackage[hyphens]{url}  
\usepackage{graphicx} 
\usepackage{natbib}  
\usepackage{caption} 
\DeclareCaptionStyle{ruled}{labelfont=normalfont,labelsep=colon,strut=off} 
\usepackage{algorithm}
\usepackage{algorithmic}
\usepackage{multirow}
\usepackage{float}
\usepackage{bibentry}
\usepackage{xcolor}
\usepackage{subcaption}
\newcommand{\arnote}[1]{\textcolor{green}{AR: #1}}
\newcommand{\sjnote}[1]{\textcolor{cyan}{SJ: #1}}

\newcommand{\high}[1]{\textcolor{red}{#1}}

\newcommand{\correction}[1]{\textcolor{blue}{#1}}

\usepackage{newfloat}
\usepackage{listings}
\floatstyle{ruled}
\newfloat{listing}{tb}{lst}{}
\floatname{listing}{Listing}
\title{A Deep Dive into the Disparity of Word Error Rates\\ Across Thousands of NPTEL MOOC Videos}

\author{Anand Kumar Rai, Siddharth D Jaiswal, Animesh Mukherjee}
\affiliations{Indian Institute of Technology, Kharagpur, India}

\begin{document}

\maketitle

\begin{abstract}
Automatic speech recognition (ASR) systems are designed to transcribe spoken language into written text and find utility in a variety of applications including voice assistants and transcription services. However, it has been observed that state-of-the-art ASR systems which deliver impressive benchmark results, struggle with speakers of certain regions or demographics due to variation in their speech properties. In this work, we describe the curation of a massive speech dataset of 8740 hours consisting of $\sim9.8$K technical lectures in the English language along with their transcripts delivered by instructors representing various parts of Indian demography. The dataset is sourced from the very popular NPTEL MOOC platform. We use the curated dataset to measure the existing disparity in YouTube Automatic Captions and OpenAI Whisper model performance across the diverse demographic traits of speakers in India. While there exists disparity due to gender, native region, age and speech rate of speakers, disparity based on caste is non-existent. We also observe statistically significant disparity across the disciplines of the lectures. These results indicate the need of more inclusive and robust ASR systems and more representational datasets for disparity evaluation in them.\footnote{\high{This work has been accepted for publication at ICWSM 2024}}
\end{abstract}

\section{Introduction}

Automatic speech recognition (ASR) systems have become increasingly prevalent in recent years with applications including voice assistants, transcription services and language translation. Auto-generated transcripts serve an integral part in providing equitable access of online video content to a wide variety of individuals and groups while voice based assistants enable users to avail a lot of online services with voice-based commands. In the past two decades, designing efficient ASRs have been an active area of research resulting in substantial advancement in the accuracy of these tools~\cite{hannun2021history}.

\subsection{Learning through video lectures}

The advent of COVID-19 pandemic has hastened the pace of adoption of online education and, Massive Open Online Courses (MOOC) platforms like NPTEL~\cite{krishnan2009nptel}, Coursera~\cite{coursera} etc. play a pivotal role in it. 
The transcripts of these videos are typically used to generate captions~\cite{kent2018case} for the videos. Such captions reduce the disconnect between the viewer and speakers with distinct accents and speaking styles. Many speakers rely on ASR tools like YouTube~\cite{Youtube_Live}, Zoom~\cite{Zoom_Live}, Otter.AI ~\cite{otterai}, etc. for generating automated transcriptions of their videos to reduce their own manual effort. Thus the error from these ASRs can directly impact the video captions and hence the overall understanding of the viewer.

\subsection{Disparity in caption generation}

These platforms are expected to work without disparity for all speaking rates, accents, tonality, independent of gender or age to ensure the generation of correct transcripts, and therefore captions, so that the listener does not misinterpret the speaker. However, concerns have been raised about the potential for such systems to exhibit bias ~\cite{feng2021quantifying,koenecke2020racial,tatman2017gender} toward certain demographics.

There has been ongoing research evaluating disparities in ASRs toward different racial~\cite{koenecke2020racial}, gender~\cite{tatman2017gender} and other groups~\cite{feng2021quantifying}. 
Most of these studies have focused on Western languages, accents, and social demographic groups. There have also been efforts to create public datasets like Artie Bias~\cite{meyer2020artie}, CORAAL~\cite{CORAAL}, AAVE~\cite{AAVE}, TED-LIUM~\cite{rousseau2012ted}, Librispeech~\cite{panayotov2015librispeech} etc., with dialectical and vernacular variations to evaluate the performance of existing ASR platforms, but these too have primarily focused on Western speakers. Moreover, there has been a lack of datasets exploring technical or pedagogical content.

\if{0}
\correction{

These platforms are expected to work without disparity for all speaking rates, accents, tonality, independent of gender or age to ensure correct transcripts are generated and the listener does not misinterpret the speaker. However, concerns have been raised about the potential for such systems to exhibit bias ~\cite{feng2021quantifying,koenecke2020racial,tatman2017gender} towards certain demographics.
}

\correction{
There has been ongoing research evaluating disparities in ASRs toward different racial ~\cite{koenecke2020racial}, gender~\cite{tatman2017gender} and other groups ~\cite{feng2021quantifying}. 
Most of these studies have focused on Western languages, accents, and social demographic groups. There have also been efforts to create public datasets like Artie Bias~\cite{meyer2020artie}, CORAAL~\cite{CORAAL}, AAVE~\cite{AAVE}, TED-LIUM~\cite{rousseau2012ted}, Librispeech~\cite{panayotov2015librispeech} etc., with dialectical and vernacular variations to evaluate the performance of existing ASR platforms, but these too have primarily focused on Western speakers. Moreover, there has been a lack of datasets exploring technical or pedagogical content. 
}
\fi
\subsection{ASR disparity impact in Indian context}

The problem of disparity in ASR systems is of particular concern in the Indian context, given the country's highly diverse linguistic landscape across its vast geography. With the rapid advancement in digital economy in the country, it has become the second largest country in terms of internet users\footnote{https://www.statista.com/statistics/271411/number-of-internet-users-in-selected-countries/} with around 0.9 billion users.Moreover, a significant number of students within India consume educational content from YouTube or other e-learning platforms~\cite{buddayya2019benefits}, but there have been no studies evaluating the transcription accuracy of ASRs for educational content delivered by Indian speakers. Due to the scale at which content is disseminated and consumed in India, it is important for researchers to study the disparities in widely deployed and state-of-the-art ASRs to ensure proper learning pedagogy is maintained.

\subsection{Our contributions}
In this study, we evaluate the correctness of captions for two popular ASR platforms -- YouTube Automatic Captions ~\cite{Youtube_Live} and OpenAI's Whisper ~\cite{radford2022robust} on a new large-scale dataset developed by us. This dataset, developed by us and called Technical Indian English (TIE), has more than 9800 educational videos from the NPTEL MOOC learning platform, comprising over 8700 hours of content and 62 million spoken words. These lecture videos are delivered by 332 speakers belonging to different gender, caste, age and regions within India. Ours is the first such annotated dataset on Indian educational videos and our in-depth analysis of the performance of the two ASRs on this dataset shows that disparities exist not only between the two platforms but also between categories of speakers such as gender and experience level on each platform. While studies have been conducted to evaluate YouTube's ASR capabilities, ours is the first study to evaluate OpenAI's Whisper for its accuracy on technical English content, spoken by non-native English speakers.

We now list the research questions that we address in this study.

\noindent \textbf{RQ1.} Since no prior dataset exists for studying ASR disparities across educational videos for non-native speakers the first question we were faced with was to develop such a dataset from scratch. In response we built the TIE dataset.

\noindent \textbf{RQ2.} How do the two ASRs perform on the TIE dataset? We perform statistical tests to identify which of the two platforms is better at transcribing technical educational content spoken by Indian speakers.

\noindent \textbf{RQ3.} Do the two ASR platforms exhibit disparity in performance among the different categories within each attribute like gender, caste, age, etc. Here as well we perform statistical tests to identify the disparities in performance not only for a given platform but also across platforms.

Finally, we list our contributions. In this paper, we propose the TIE dataset of 9860 audio files spanning a massive duration of 8740 hours worth of lectures sourced from the NPTEL MOOC platform and annotated for attributes corresponding to the lectures and the 332 instructors who have delivered these lectures. We believe that this dataset can serve as an excellent benchmark for studying the performance of various AI softwares including ASRs. We audit the YouTube Automatic Captions and OpenAI Whisper ASRs for existing disparities towards gender, caste, teaching experience, native region of the speakers, speech rate and technical discipline of the lectures. 

Our analysis reveals a significant disparity against non-native English speakers in the Indian population,  particularly for male speakers, speakers from southern India and those with a slow speech rate. In addition, we found that the experience of speakers and the topic of speech had an impact on the accuracy of ASRs. We also observed that while Whisper had better overall accuracy than YouTube, disparities were higher in Whisper. We conclude by discussing possible causes of these disparities and proposing strategies to address them.

\if{0}
This digitally enabled population uses a variety of online applications like social media, e-commerce, video-sharing platforms etc as a daily life routine and by analyzing the performance of two ASR systems namely YouTube Automatic Captions and OpenAI Whsiper on a large and diverse dataset of Indian speakers, we hope to shed light on any patterns of systematic error or disadvantage that may exist in the voice assistance and transcription services due to these models.

To understand different kinds of bias towards linguistically diverse and acoustically varied sections of society in transcribing and recognizing their voices present in the state-of-art commercial ASR systems, various publicly available datasets ~\cite{meyer2020artie,CORAAL,AAVE}  are used. However none of the existing public dataset having representation from varied English accent and speech style of Indian speakers in a pedagogical context.

In this work, we are proposing a self-curated Technical Indian English(TIE) dataset of 9860 YouTube hosted NPTEL lecture videos from 23 science and engineering disciplines spanning a total time duration of 8740 hours having 62.2 M word utterances and delivered by 332 instructors.

The dataset consists of audio files and with its ground-truth transcripts and demographic attributes of speaker like gender, caste category, teaching experience and native region annotated.The speech rate attribute derived from audio duration and manual transcript word length , lecture topic and  associated discipline group  corresponding to each audio file is also labelled.

In this work, we have used our TIE dataset to perform  the transcription error disparity analysis of two very relevant ASR platforms viz. YouTube Automatic Captions ~\cite{Youtube_Live} and  Whisper ~\cite{radford2022robust}  with respect to speaker and speech characteristics.The analysis will be focussed on following three research questions:
\begin{itemize}
    \item Do gender and caste group based social disparities which is prevalent in Indian demography gets reflected in performance of ASR systems under investigation ?
    \item Do speech related characteristics like accent, pitch and speech rate of Indian speakers affects the transcription errors ?
    \item How good are  ASRs in generalizing for technical domain specific datasets ?
\end{itemize}

 The results of our analysis show the presence of significant disparity due to demographic characteristics of instructors. We have also discovered that the transcription is much better for instructors  who speak faster than for slow speakers and for the lectures belonging non-engineering category than for engineering discipline lectures . We conclude the paper by discussing possible reasons that explain the disparities we measured, and possible mitigation strategies.

To support future research, upon paper publication we will publicly release on Github the following:

\begin{itemize}
\item Our new corpus TIE speech dataset along with ground truth transcripts and annotated metadata related to the audio files ( 9860 files,62.2 M words, 8740 hours, 332 speakers).
 \item A total of 19720 transcripts that we have generated for our experiments using from YouTube and Whsiper ASR.
 
\end{itemize}

I
 \fi
\if{0}

This work of ours is focused on investigating the following research questions:
\begin{itemize}
\item	Is performance of Acoustic Model of ASR systems fair across various demographic groups in Indian context?

\item Is performance of Language Model of ASR systems fair across disciplines of engineering education?
\end{itemize}
We have evaluated two ASRs, viz YouTube Captions and Open AI Whisper as a part of our investigation to answer the above-mentioned questions. YouTube Captions is being used at scale for auto transcriptions of videos by YouTube while Open AI Whisper Model is an open-source model which has been released recently and claimed to have best performance across datasets. There has been various works dedicated to the analysis of YouTube Captions ASR model but none of them has been dedicated analysing its performance on the Indian voices. Also, on comparing with Open-Source Model like Whisper, it will be possible to comment on the accuracy, reliability and robustness with state-of-the-art Whisper’s ASR Language and Acoustic Model.

The dataset we have used in this work is a self-curated dataset of 9860 YouTube hosted NPTEL [9] lecture videos from 23 science and engineering disciplines spanning a total time duration of 8740 hours and delivered by 332 instructors. There are several publicly available datasets that are commonly used for evaluating ASR systems, such as the Wall Street Journal Corpus, the TED-LIUM Corpus, and the VoxCeleb Corpus. These datasets contain transcriptions of audio recordings of spoken language and are used to evaluate the performance of ASR systems on tasks such as transcription, translation, and language identification. However, we have used the self-curated because none of the public datasets has such large-scale transcriptions of varied Indian accents which enables to test robustness of acoustic model of ASR systems from Indian context. In addition, these transcriptions are from 23 different engineering domains from Aerospace to Agriculture, which will help in evaluating the Language Model of ASRs under investigation.

Our results show that overall Word Error Rate (WER) of Whisper is lower than that of YouTube captions ASR but there is significant disparity in WER across demographic groups and engineering disciplines in both ASRs. We conclude the paper by pointing out the possible reasons to explain the overall difference in accuracy of both ASRs and the bias noted in acoustic and language model of both ASRs. 

\sjnote{Heavy re-write required. Currently, flow of thought and concepts is missing. Need to introduce concepts first before building upon them to explain the main gist of the paper. language/acoustic models are not defined. use cases are not explained with citations etc.}

\fi
\section {Related Work}

\subsection{Automatic speech recognition}
ASR systems, designed to recognize and translate spoken language have been actively developed since the 1950s. Today, this technology has become highly ubiquitous ~\cite{juniperresearch_number_of_voice_assistant_devices} and has various use cases ranging from generating captions for pre-recorded videos~\cite{Youtube_Live} and live videos~\cite{Zoom_Live} to personal voice assistants like Siri~\cite{Siri_Live}, Alexa~\cite{alexa} and Cortana~\cite{Cortana_Live}. 


\subsection{Audit of ASRs}
Recently, there has been a growing interest in auditing~\cite{ngueajio2022hey} these ASR platforms for potential disparities against various gender~\cite{adda2005speech,tatman2017effects,tatman2017gender,garnerin2019gender} and racial groups~\cite{koenecke2020racial,tatman2017effects}. \cite{koenecke2020racial} have highlighted the existing racial bias against Black speakers prevalent in state-of-the-art commercial ASRs by benchmarking them against CORAAL~\cite{CORAAL} and AAVE~\cite{AAVE} datasets having high representation of speakers from African American community.~\cite{adda2005speech,garnerin2019gender} have pointed out the gender bias in ASR systems which favors female speakers when they benchmarked the ASRs performance for English and French news broadcast dataset.~\cite{vipperla2010ageing} have audited the impact of speaker age on the ASR performance. Most of the existing studies fall under the purview of \textit{black box audits}~\cite{sandvig2014auditing} due to the lack of access to model architecture and training data for ASRs supplied by commercial vendors~\cite{koenecke2020racial,tatman2017effects}.

\subsection{Non-native English datasets and models} 
The authors in~\cite{meyer2020artie} proposed a dataset which had speaker age, accent and gender annotated along with the their associated audio and benchmarked the DeepSpeech model~\cite{hannun2014deep}. They reported more accurate transcriptions of US English compared to Indian English. Some of the other curated datasets to facilitate benchmarking of ASRs on non-native speakers are L2-Arctic~\cite{zhao2018l2}, 
EdAcc~\cite{sanabria2023edinburgh} and AccentDB~\cite{ahamad2020accentdb}. 

In addition~\cite{hinsvark2021_asrsurvey} presented a comprehensive survey on bias in ASR systems due to variety in the speakers' accents. ~\cite{sullivan2022improving},~\cite{shibano2021speech} and~\cite{vu2014improving} presented different approaches like transfer learning and language model decoding that allow ASRs to perform better on non-native speaker dataset.

\subsubsection{The present work.} In order to bridge the lack of data we present a large-scale dataset of technical lecture videos comprising 8740 hours of speech by 332 Indian speakers on more than 20 diverse lecture topics. Each video file has been annotated with demographic attributes like teaching experience, gender, caste and native region of respective speaker plus audio metadata like speech rate, discipline and topic of the lectures. Unlike~\cite{meyer2020artie}, our dataset is highly diverse comprising speakers from all the four regions of India. In addition, our dataset is richer than AccentDB~\cite{ahamad2020accentdb}, another Indian ASR dataset in terms of the number and diversity of native speakers represented and the total duration of speech data available. In particular, while AccentDB includes only speakers with native languages such as Bangla, Malayalam, Odiya and Telugu and has a duration of only 9 hours, our dataset covers a wider range of Indian languages and comprises a total duration of 8740 hours. 
Next, existing literature~\cite{tatman2017effects,tatman2017gender} has focused on US speakers' whose native language is English. Ours is one of the first \textit{large-scale} study focusing on non-native English speakers from the Global South. Finally, ours is the first study to perform a comprehensive audit of OpenAI Whisper on technical speech by non-native English speakers.

\if{0}
In this work, we are doing three very important contributions to the existing bias literature on ASR systems.First, we are proposing a new Technical Indian English  (TIE dataset) which we will release under a Creative Commons CC0 license – the most open and permissive license for data. This is a curated subset of the 8740 hours lectures from 332 Indian instructors on 23 diverse engineering and non-engineering domain from the NPTEL lecture series.The audio files have been annotated with demographic attributes like teaching experience, gender, caste and native region of respective speaker and metadata of the audio like speech rate, discipline group , discipline and topic of lecture. This dataset has labelled regional variation of Indian English which is not present in ~\cite{meyer2020artie} and can act as an intersectional benchmark dataset for auditing both speaker related and content related bias in ASR systems.

Secondly we have used the proposed dataset to understand how the performance of YouTube Automatic Captions ASR fared across aspects of Indian pedagogical videos.As works like ~\cite{tatman2017effects,tatman2017gender} has analyzed the YouTube Automatic Captions for gender, race and dialect bias with respect to diversity in American speakers but the same has not been conducted in Indian context by any previous work.Also they have have audited the robustnes on YouTube Automatic Captions from phonetic and acoustic variance point of view but our work will also test its performance against lexical rich vocabulary utterances from different domains.

And lastly , we have audited the demographic and non-demographic disparity in performance of OpenAI Whisper, which has claimed human-level transcription accuracy without any domain specific additional training for their model. There are works like ~\cite{olivier2022there} and ~\cite{saadany2022better}  which has tested the robustness of Whisper model on adversarial audio samples and  domain adaptation of the model in transcribing UK Supreme Court hearings respectively. But a comprehensive audit of the model comparing its disparity in performance due to demographic traits of speakers on a large scale dataset like TIES dataset has not been attempted by any previous work.
\fi



\if{0}

~\cite{hinsvark2021_asrsurvey} has surveyed the works on bias in ASR systems due to accent of the speakers and has highlighted  the English accent variation due to regional factors is not captured in most of the dataset used by various works.
While the studies mentioned above highlighted the disparity in ASR performance due to acoustics and phonetic variation present in speaker voice , .

Our study will add an important and unexplored perspective to the existing literature as we are undertaking bias study of two ASR systems , Youtube Captions ASR which is being used at a mass commercial scale and the latest open-source OpenAI Whisper which promises human level transcription accuracy on the newly curated TIE dataset which encompasses attributes of Indian demography.

of stressed upon the requirement and methodology for creating better bench-marking datasets for ASR systems   and these works have inspired us to curate a new dataset, Technical Indian English(TIE) covering  Technical educational lectures in diverse Indian English accent and demographical attributes of Indian society.

\cite{garnerin-etal-2021-investigating} has concluded that a more balanced representational training dataset in terms of auditory properties and gender respectively will result in lower disparity towards the targeted group.

~\cite{aksenova2021_benchmark} has stressed upon the requirement and methodology for creating better bench-marking datasets for ASR systems

Studies have been performed that have identified gender biases in ASRs against male speakers~\cite{adda2005speech,tatman2017effects,tatman2017gender,garnerin2019gender} and racial bias against black speakers~\cite{koenecke2020racial,tatman2017effects} in state-of-art commercial ASRs. Researchers~\cite{hinsvark2021_asrsurvey,tatman2017effects,tatman2017gender} have also identified that native accent and dialect has an impact on the ASR's auto-generated transcript's accuracy. While the previous studies ahve highlighted the disparity in ASR performance due to variation in acoustics in speaker voice,~\cite{goldwater2010_words,markl2021context,282844222}  have further broadened the spectrum of bias literature in ASR system by studying the bias due to content diversity in speech.

As has been often suggested in fairness literature, bias in performance could be a result of an imbalanced dataset and most of the previous works have advocated for the use of more diverse and balanced dataset in terms of demographic and linguistic context. ~\cite{Ko2015AudioAF} and \cite{open_speech} showed that a more balanced representational dataset in terms of auditory properties and gender respectively will result in lower disparity towards the targeted group. ~\cite{aksenova2021_benchmark} stressed upon the requirement and methodology for creating better bench-marking datasets for ASR systems.

\fi
\section{Dataset and Platforms}
In this section, we give a detailed overview of the dataset that we curate for this audit study, along with a description of the platforms that we evaluate this dataset on. 

\subsection{TIE dataset}

\begin{table*}[!ht]
\scriptsize
    \centering
\begin{tabular}{|cc|c|c|c|c|}
\hline
\multicolumn{1}{|c|}{\textbf{Attribute}} & \textbf{Category} & \textbf{\% Speakers} & \textbf{\% Lectures} & \textbf{\# Hours} & \multicolumn{1}{l|}{\textbf{\begin{tabular}[c]{@{}l@{}}\#Words in ground \\ truth transcripts\end{tabular}}} \\ \hline
\multicolumn{1}{|c|}{\multirow{2}{*}{Gender}} & Female & 5.4 & 5.8 & 498 & 3.7 M \\ \cline{2-6} 
\multicolumn{1}{|c|}{} & Male & 94.6 & 94.2 & 8242 & 58.5 M \\ \hline\hline
\multicolumn{1}{|c|}{\multirow{2}{*}{Caste}} & Res. & 27.4 & 27.0 & 2285 & 16.2 M \\ \cline{2-6} 
\multicolumn{1}{|c|}{} & Unres. & 72.6 & 73.0 & 6455 & 46.0 M \\ \hline\hline
\multicolumn{1}{|c|}{\multirow{4}{*}{Experience}} & $\leq 1980$ & 14.5 & 14.2 & 1287 & 9.0 M \\ \cline{2-6} 
\multicolumn{1}{|c|}{} & 1981-90 & 22.9 & 22.1 & 1895 & 13.1 M \\ \cline{2-6} 
\multicolumn{1}{|c|}{} & 1991-00 & 32.8 & 33.9 & 3033 & 21.7 M \\ \cline{2-6} 
\multicolumn{1}{|c|}{} & $\geq 2001 $ & 29.8 & 29.8 & 2526 & 18.5 M \\ \hline\hline
\multicolumn{1}{|c|}{\multirow{4}{*}{Native Region}} & East & 35.2 & 37.9 & 3405 & 23.1 M \\ \cline{2-6} 
\multicolumn{1}{|c|}{} & West & 8.4 & 6.6 & 584 & 4.3 M \\ \cline{2-6} 
\multicolumn{1}{|c|}{} & North & 21.7 & 19.2 & 1663 & 12.3 M \\ \cline{2-6} 
\multicolumn{1}{|c|}{} & South & 34.7 & 26.3 & 3088 & 22.5 M \\ \hline\hline
\multicolumn{1}{|c|}{\multirow{3}{*}{Speech Rate}} & Slow & 72.9 & 39.1 & 3442 & 19.3 M \\ \cline{2-6} 
\multicolumn{1}{|c|}{} & Average & 71.4 & 21.4 & 1880 & 13.3 M \\ \cline{2-6} 
\multicolumn{1}{|c|}{} & Fast & 75.3 & 39.5 & 3418 & 29.6 M \\ \hline\hline
\multicolumn{1}{|c|}{\multirow{2}{*}{Discipline}} & Engg. & 70.5 & 70.8 & 6269 & 44.8 M \\ \cline{2-6} 
\multicolumn{1}{|c|}{} & Non-engg. & 30.7 & 29.2 & 2472 & 17.4 M \\ \hline
\multicolumn{4}{|c|}{\textbf{TOTAL}}  & \textbf{8740} & \textbf{62.2 M} \\ \hline
\end{tabular}
    \caption{Metadata statistics for the Technical Indian English (TIE) dataset. The attributes and the categories therein, are presented in the rows. \% Speakers and \% Lectures represent the share of speakers and lectures from 332 speakers and 9860 lectures respectively. \# Hours and \# Words represent the time and number of words (in Millions) for all lecture videos corresponding to a given attribute's category. There are approximately 30 lectures per speaker, and each lecture runs for $\sim 53$ minutes and has $\approx$ 6300 words.}
    \label{tab:dataset_metadata}
\end{table*}

In this study, we curate a new large-scale dataset\footnote{\high{Available at https://github.com/raianand1991/TIE/}} of 9,860 lecture videos from the NPTEL (National Programme on Technology Enhanced Learning)~\cite{krishnan2009nptel} platform, which is a government-funded joint initiative of the Indian Institutes of Technology (IITs)\footnote{https://en.wikipedia.org/wiki/Indian\_Institutes\_of\_Technology} and the Indian Institute of Science (IISc)\footnote{https://en.wikipedia.org/wiki/Indian\_Institute\_of\_Science} and provides high-quality educational content in the form of video lectures, online courses and other resources to students and educators throughout India. The NPTEL platform has more than 2500 courses across 29 engineering and non-engineering disciplines with over 78,000 videos distributed on both the NPTEL website\footnote{https://nptel.ac.in/courses\label{nptel}} as well as YouTube. The official YouTube channel of NPTEL\footnote{https://www.youtube.com/@iit} has more than 20,000 videos from 22 disciplines with more than 2 million subscribers and around 400 million cumulative views on all videos. The platform is similar to other MOOC platforms like MIT OCW~\cite{ocw2008opencourseware}, Coursera~\cite{coursera}, edX ~\cite{edxFreeOnline} etc.

For this study, we sample 9860 videos from NPTEL's YouTube channel for which both ground truth transcripts and YouTube captions are available in English. These lecture videos have been delivered by 332 speakers of Indian origin belonging to both genders -- male \& female, reserved and unreserved caste groups~\cite{caste_definition}, multiple  experience ranges and residing in diverse geographical regions. The speakers are faculty members at the premier educational institutes of India (e.g., IITs and IISc). Thus our dataset is both acoustically and linguistically rich and diverse. Each lecture video is $\approx 53$ minutes long, giving us a massive 8740 hours of lecture videos with more than 62 million spoken words in total. We note that even though each individual speaker speaks in $\approx 31$ videos, there are variations in the video content, speaker's tone, accent, and style. Hence each video lecture can be considered to be a unique input video resulting in 9860 unique data points. 

All the video lectures collected as part of this dataset are non-interactive and pre-recorded technical monologues delivered by the lecturers. Such monologues are generally less noisy than lectures delivered in interactive settings. This also helps us in avoiding speech preprocessing and enhancement tasks which could potentially add bias to the dataset. More details regarding the dataset metadata are available in Table~\ref{tab:dataset_metadata}. 

\subsection{Dataset preparation}
We prepare the TIE dataset by sampling a list of courses from the NPTEL ~\cite{krishnan2009nptel} website in two phases. In the first phase, we collect metadata for all courses -- name, discipline, institute, instructor and the course weblink. This resulted in a corpus of 2567 courses. Each course has $\approx 31$ lecture videos, for which we collected the titles and weblinks. This resulted in a total dataset size of 78,222 lecture videos. In the second phase, we filter out only those videos which were hosted on YouTube and had ground truth transcripts available. This filtered dataset has 9860 lecture videos along with all metadata information mentioned previously. We collect all video files and extract the audio tracks in MP3 format. The size of the final corpus of 9860 audio files is approximately 700 GB.




\subsection{Dataset annotation}
We annotate each video file in our dataset with multiple features for the speaker viz. gender, caste, experience, lecture discipline, affiliation and native region. Here, we define experience as the year since which the speaker has been affiliated to their institute and the native region as the geographical region to which the speaker belongs -- north, east, south or west India. We do the above by scraping the NPTEL course website to collect information regarding the course, discipline, institution, instructor and the Youtube URL for the videos.

\subsubsection{Demographic information annotation}
Next, to identify the demographic information of the speaker, we use a pre-trained BERT-based model proposed by~\cite{Medidoddi2022_decoding}, which takes as input the individual's full name and returns the gender and caste for the same. This model has a claimed accuracy of 96.06\% for gender classification and 74.7\% accuracy for caste category classification. Note that our definition of gender represents the perceived gender, and not necessarily the self-identified gender of the speaker. Similarly, the caste of individual speakers represents the perceived caste category which is generally determined by the surname of the individual. We have used only two categories for caste -- unreserved (historically privileged groups) and reserved (SC/ST/OBC -- historically discriminated groups) for ease of analysis.In order to check the robustness of the accuracy of the caste classification model reported in~\cite{Medidoddi2022_decoding} we manually verified its performance for a random sample of 50 speakers from our TIE dataset. We obtained an accuracy of 80\% in identifying speaker caste category based on their names for our dataset.


For annotating the experience of speakers, we use the year when they started their teaching career.  This information is collected from the speakers' curriculum vitae (CV) or profile page. We created four categories for experience viz. $\leq 1980$, 1981-90, 1991-2000, and $\geq 2001$ indicating the time range when the speaker started their professional career.

To annotate for the native region, we use the information available in the speaker's CV. If this is unavailable, the surname of the speaker is used to infer the native state as Indian names are based on naming conventions that are region-specific\footnote{https://en.wikipedia.org/wiki/Indian\_name}. The inter-annotator agreement (Cohen's $\kappa$ = 0.71) of co-authors is used for finalizing the annotation for the native state attribute. 

\subsubsection{Non-demographic information annotation}
Finally, we also annotate the dataset with non-demographic information like speech rate and the speaker's affiliation and topic discipline (engineering or non-engineering). Speech rate is a measure of the word utterance rate of the speaker and is calculated in terms of words per minute (wpm). We calculate this value by dividing the total number of words in the ground truth transcript by the lecture duration. We use this speech rate to categorize the dataset into three classes viz. slow, average and fast representing those with $\leq$ 33, 33-66, and $\geq$ 66 percentile speech rates respectively. 

\subsubsection{Dataset statistics}
Table~\ref{tab:dataset_metadata} presents the statistics for the TIE dataset that we curate as part of this study. We can immediately see that there is a significant skew toward male speakers ($\approx 95\%$) and those belonging to the unreserved caste category ($\approx 73\%$). This is representative of the distribution observed in higher educational institutes in India. ~\cite{prathap2017excellence,chanana2006gender} have reported on the marginalized representation of females in teaching and leadership profiles in Indian Higher Education institutions and in fact a recent survey\footnote{https://aishe.gov.in/} from the year 2019-20 amongst Institutes of National Importance in India shows that females occupy only 20\% of the teaching positions. Similarly, it has been observed that only 21\% of the lecturers in higher education institutes belong to the reserved caste category. In terms of teaching experience, the dataset has a higher representation ($\approx 63\%$) of speakers who started teaching after 1991. The distribution of speakers from the native regions is fairly balanced except for west India which is underrepresented with a share of only 8.4\% speakers. We annotate speech rate and the discipline for each lecture, instead of each speaker. Hence, there is an overlap of speakers across these subcategories resulting in the total sum exceeding 100\%. In the speech rate category, it can be seen that most lectures are delivered by speakers speaking either slowly or fast (each having a 39\% share). As  NPTEL hosts lectures from higher educational institutes specializing in engineering, the share of engineering lectures has a majority (70.8\%).

\subsection{Platforms evaluated}
In this study, we audit two ASR platforms -- YouTube Automatic Captions~\cite{Youtube_Live} and OpenAI's Whisper~\cite{radford2022robust}. While Youtube is a full-fledged video-sharing platform with its own proprietary ASR solution to generate captions for uploaded videos, Whisper is an open-source ASR system developed by OpenAI that can generate transcripts in multiple languages. 

We now give a brief description of the two platforms under consideration.

\begin{itemize}
    \item \textsc{Youtube automatic captions}: YouTube launched its proprietary ASR software -- YouTube Automatic Captions in 2009 to generate automatic captions for videos uploaded on YouTube. The captions are textual representations of the audio and get displayed on the screen while the video is being played. The feature was developed to make it easier for people with hearing impairments to access the video content on the platform. As this is a commercial black-box model, the architecture and training data are not available in the public domain.
    
    \item \textsc{Whisper}: This is an open-source ASR system launched by OpenAI in 2022. The architecture is based on an end-to-end encoder-decoder transformer and trained on 680k hours of multilingual and multitask supervised data collected from the web. The authors claim that their model has improved robustness to accents and technical language. The model is available in nine variants of multiple sizes viz. tiny, base, small, medium, and large. The model architecture has been open-sourced and is available for zero-shot transcription of speech audio in multiple languages.   
\end{itemize}

\noindent\textit{Reasons for choosing these platforms}: We use the above two ASR platforms for disparity evaluation due to the following reasons.
 \begin{itemize}
  
     \item  YouTube ASR is directly used for generating captions for the YouTube videos, which is by far the largest commercial video-sharing platform. Thus the transcripts used for caption generation has a significant influence on the accessibility and reach of the videos on the platform. The NPTEL videos used in our TIE dataset have more than 200M views cumulatively along with 831K likes on YouTube, indicating wide content consumption. 
     \item  
     
     We experimentally observe that Whisper outperforms other open-source ASRs like DeepSpeech~\cite{hannun2014deep} and Wav2Vec 2.0~\cite{baevski2020wav2vec} in terms of overall WER on the TIE dataset when evaluated in a zero-shot setting. For this evaluation, we use pre-trained DeepSpeech\footnote{https://deepspeech.readthedocs.io/en/r0.9/USING.html}, Wav2Vec 2.0\footnote{https://huggingface.co/facebook/wav2vec2-large-robust-ft-libri-960h} and Whisper\footnote{https://huggingface.co/openai/whisper-base.en} models trained on English datasets.  The median WER on TIE dataset for these models are 96.4\%, 28.6\% and 11.8\% respectively. The performance of Whisper being substantially better than the other two competing SOTA models, we use it for all our subsequent experiments.
 \end{itemize}
\section{Methodology \& Evaluation Metrics}
We now describe the experimental methodology and the metrics used to evaluate the observations. We first discuss the process of collecting the ASR-generated transcripts, followed by a description of the metric -- word error rate (WER) and the statistical tests performed to determine the disparity among the various categories for each attribute as well as between the two ASRs under consideration.

\subsection{Methodology}
We generate the transcripts for each video in our TIE dataset using the transcription services of both the platforms-- YouTube and Whisper. For YouTube, we use the \textsc{YoutubeTranscript} API~\cite{pypiYoutubetranscriptapi} to generate the textual transcripts and post-process these by removing frame-wise timestamps. To generate transcripts for Whisper, we first extract the audio in MP3 format and provide these as input to the pre-trained base model of Whisper (base.en), trained on an English corpus. Each MP3 file is of approx 50 min duration and inference on Whisper base model took $\approx 2$ mins per file when executed on a server with 2 Tesla P100 PCIe 16 GB GPUs. The overall inference time in generating transcripts from Whisper for the entire dataset of 9860 files was $\approx 250$ hours.

\subsection{Evaluation metric}

We choose WER as the evaluation metric as it is the most commonly used metric for evaluating ASRs, having been used by many studies in literature~\cite{tatman2017effects,tatman2017gender,koenecke2020racial,garnerin2019gender} as well as for benchmarking public datasets. The WER is measured in two steps -- (i) a preprocessing step in which we remove all the punctuation marks and standardize the case of all words and (ii) the WER calculation step for which we use the JiWER~\cite{Jiwer_LIB} Python library to compare the ground truth and ASR generated transcripts.

In addition to being the most popular measure, WER  also presents a better idea of what went wrong in the automatic transcription. It does so by taking into account all types of errors, including insertion, deletion, and substitution errors as WER is defined as  \[WER =  \frac{S+D+I}{N}  \] where $S$, $D$ and $I$ are the number of substitutions, deletions and insertions respectively needed to transform the reference text into the hypothesis text and $N$ is the number of words in the reference text. The share of these three components in WER can explain the underlying cause of the error.
    
In order to further strengthen our observations we also compute two more recently introduced measures BERTScore~\cite{tobin2022assessing} and SemDist~\cite{kim2021semantic} that focus on semantic closeness of ground-truth sentences and ASR generated sentences. 

%

\if{0}We choose WER as the evaluation metric for the following reasons.
\begin{itemize}
    \item Wide Usage: WER is the most commonly used metric for evaluating ASRs, having been used by many studies in literature~\cite{tatman2017effects,tatman2017gender,koenecke2020racial,garnerin2019gender} as well as for benchmarking public datasets. 
    \item We evaluated both Youtube and Whisper ASRs using SemDist and BERTScore. The median SemDist score for Youtube was 0.038 and for Whisper was 0.01. Similarly, the BERTScore for Youtube was 0.843 and for Whisper was 0.891. Thus Whisper's transcripts have a higher level of human-like understanding compared to Youtube but these measures do not give any insights into the performance difference between the two platforms. 
    \item WER is sensitive to errors: It takes into account all types of errors, including insertion, deletion, and substitution errors as WER is defined as  \[WER =  \frac{S+D+I}{N}  \], where S,D and I are the number of substitutions, deletions and insertions needed to transform the reference text into the hypothesis text, while N is the number of words in the reference text.The share of these three components in  WER can reveal the underlying cause of the error. 
    \sjnote{can add to previous one or remove?}\arnote{Removed}
\end{itemize}
\fi




\if{0}
There are several commonly used metrics like Word Error Rate(WER) , Character error rate (CER)  etc. Some of the recent works like ~\cite{tobin2022assessing} and ~\cite{kim2021semantic} have proposed BERTScore and SemDist measures focusing on semantic closeness of ground-truth sentences and ASR generated sentences.However, we have chosen WER as an evaluation metric for this work of ours due to to the following reasons:
\begin{itemize}
    \item WER is easy to understand and interpret: It is a simple percentage that represents the number of words that are incorrectly recognized by the system.
    \item WER is widely used: It is the most commonly used metric for evaluating ASR performance, which means that there is a lot of data available for comparison purposes.Most of the bias literature in ASR ~\cite{tatman2017effects,tatman2017gender,koenecke2020racial,garnerin2019gender} have used it for reporting disparities among groups. In addition, WER is widely used as benchmarking metric by ASR models on different public datasets.  
    \item WER is sensitive to errors: It takes into account all types of errors, including insertion, deletion, and substitution errors as WER is defined as  \[WER =  \frac{S+D+I}{N}  \], where S,D and I are the number of substitutions, deletions
and insertions needed to transform the reference text into the hypothesis text, while N is the number of words in the reference text.This makes it a good metric for identifying areas where the ASR system needs to be improved.
\end{itemize}

We measure the WER in two steps, a preprocessing step  in which we alter the reference ground truth and the ASR-generated text to  WER calculation step to reduce the occurrence of false mis-transcriptions due to the presence of equivalent but different, words and a WER calculation step, in which we calculate the WER directly using the JiWER~\cite{Jiwer_LIB} Python library.
\fi

\subsection{Disparity determination}
\begin{figure}[!t]
    \centering
    \includegraphics[width= \textwidth, height=3cm, keepaspectratio]{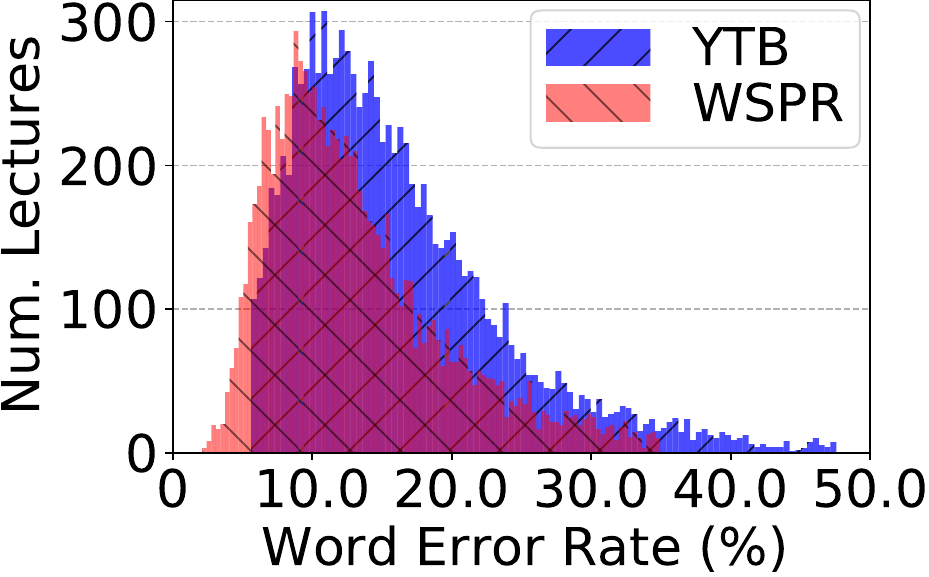}
    \caption{WER distribution for YouTube and Whisper ASRs on the TIE dataset. Both distributions are right-skewed.}
    \label{fig:histogram_wer}
\end{figure}

To measure the disparity between categories within an attribute like gender, experience, etc. we perform statistical significance tests. If the differences between the WERs for the two categories like male and female are statistically significant, we can conclude that there exists a disparity within the ASR between the two categories. From Figure~\ref{fig:histogram_wer}, we observe that the WER distributions for both the softwares are right-skewed. We, therefore, use non-parametric statistical test -- Kruskal-Wallis ($\alpha$ = 0.001) to determine whether there exist statistically significant differences between the WER medians of the various categories for every attribute. We choose a very strict $p$-value ($p<0.001$) to test the hypothesis related to disparities among all attributes to avoid any bias in our testing approach.

In addition, we also compare the median of WER components viz. insertion, deletion and substitution to ascertain the cause of disparity within each category.




\section{Results}
We now present the results from our experimental evaluation of the YouTube and Whisper ASRs' transcriptions for the TIE dataset. We first evaluate the overall performance of the two platforms, followed by an in-depth study of the disparities between the different categories for each demographic attribute and speech characteristic mentioned in Table~\ref{tab:dataset_metadata}.


\subsection{Overall disparity}
From Figure~\ref{fig:histogram_wer}, we see that for the YouTube platform, only 75.6\% videos have a WER lower than 20\% whereas this ratio is 84.0\% for Whisper's ASR transcriptions. This indicates that on average, Whisper reports lower errors in transcription compared to YouTube. 

We also evaluate both YouTube and Whisper ASRs using SemDist and BERTScore measures. The median SemDist score (the lower the better) for YouTube was 0.038 and for Whisper was 0.01. Similarly, the BERTScore (the higher the better) for YouTube was 0.843 and for Whisper was 0.891. Thus in both cases the performance of the Whisper ASR is better than the performance of the YouTube ASR which is what we also observe using the WER metric. Thus, the trends being exactly the same, to reduce verbosity we only report WER and its subparts $S, D, I$ in the rest of the analysis in the paper.


\if{0}
\begin{figure}[!t]
    \centering
    \includegraphics[width= 0.75\columnwidth, width =3cm, keepaspectratio]{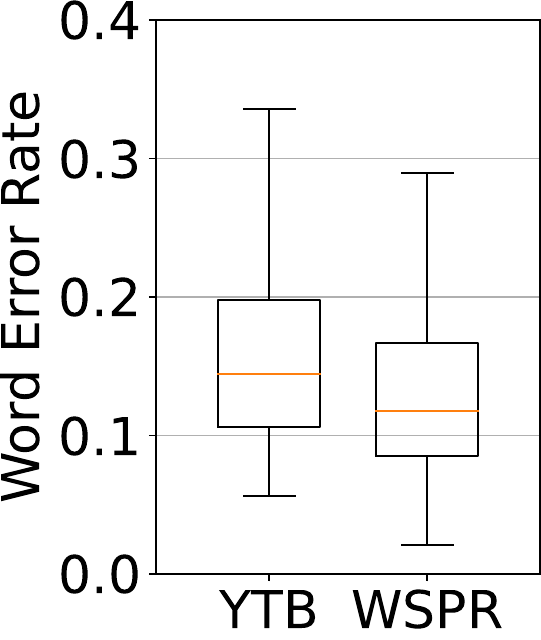}
    \caption{Boxplot distribution of the WER for Youtube (YTB) and Whisper (WSPR) ASRs on the TIE dataset. Dispersion of WER is lower for Whisper ASR from YouTube ASR both in terms of inter quartile range and min and max values.} 
    \label{fig:overall_boxplot_wer}
\end{figure}
\fi


\begin{figure}[!t]
    \centering
    \includegraphics[width= 0.75\columnwidth, height =3cm, keepaspectratio]{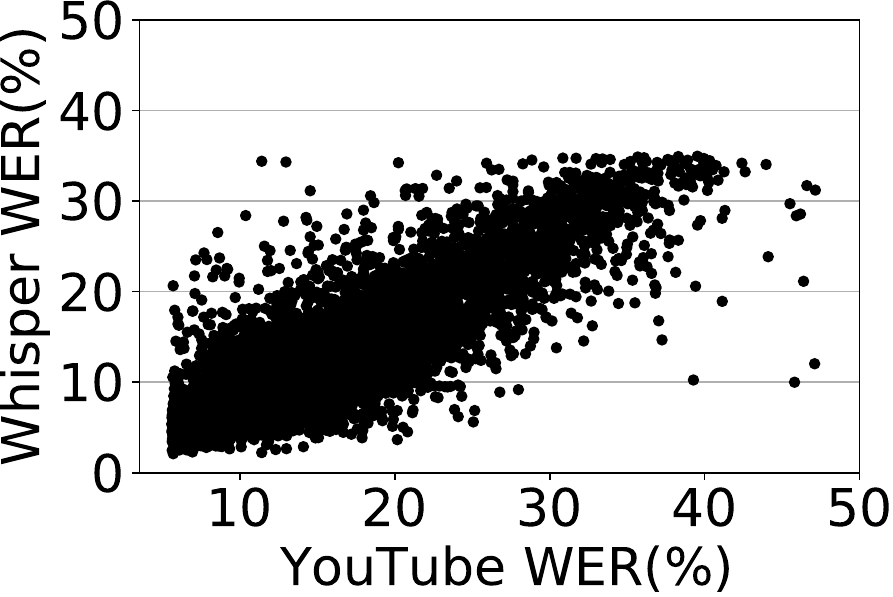}
    \caption{Correlation plot of the WER for YouTube and Whisper ASRs on the TIE dataset. The correlation co-efficient is 0.81. }
    \label{fig:correlation_wer}
\end{figure}

In Figure~\ref{fig:correlation_wer} we analyze the overall correlation between WER exhibited by YouTube and Whisper ASR corresponding to each lecture. We observe a strong positive correlation with a co-efficient of 0.81 indicating that the errors in both ASRs follow the same pattern. The lectures for which the transcription error is high in YouTube ASR-generated transcripts have a high correlation with the lectures for which the Open AI Whisper performs poorly and vice versa. However, there are few exceptions to this trend.

In Table~\ref{to_5_tbl} we report the attributes of the top \textit{five} videos with the highest WER for Whisper ASR. We observe that the majority of lectures with the highest WER are from male speakers in the unreserved caste category teaching engineering subjects. 

\begin{table}[!t]
    \centering
    \tiny
    \begin{tabular}{|c|c|c|c|c|c|}
    \hline
     \textbf{Attribute} & \multicolumn{5}{c|}{\textbf{Attribute values}} \\ 
     \hline
     \textbf{Gender} & M & M & M & M & M \\
     \hline
     \textbf{Caste} & UR & UR & UR & UR & UR \\
     \hline
     \textbf{Experience} & 1991-00 & 1981-90 & 1981-90 & \textgreater 2001 & \textgreater 2001 \\
     \hline
     \textbf{Native Region} & East & South & North & East & East \\
     \hline
     \textbf{Speech Rate} & Slow & Fast & Slow & Avg. & Fast \\
     \hline
     \textbf{Discipline} & Engg. & Engg. & Engg. & Engg. & Non-Eng. \\
     \hline
     \textbf{YouTube WER \%} & 39.5 & 38.3 & 35.7 & 36.1 & 30.9\\
     \hline
     \textbf{Whisper WER \%} & 34.9 & 34.9 & 34.9 & 34.8 & 34.7 \\
     \hline
    \end{tabular}
    \caption{Description of top 5 lectures with worst WER for Whisper ASR.  Values for the attributes of speakers corresponding to these five lectures are mentioned. It can be observed that all the five lectures with the highest WER were delivered by male speakers of unreserved caste category and the lectures were related to engineering disciplines.WER corresponding to these lectures for YouTube ASR is also mentioned for reference. M: male, UR: unreserved.}
    \label{to_5_tbl}
\end{table}

\begin{table*}[!t]
    \scriptsize
    \centering
    \begin{tabular}{|c|c|c|c|c|c|c|c|}
    \hline
     \multirow{2}{*}{\textbf{Category}} & \multirow{2}{*}{\textbf{Attribute}} & \multicolumn{3}{c|}{\textbf{YouTube}} & \multicolumn{3}{c|}{\textbf{Whisper}}\\
    \cline{3-8}
     &  & \textbf{WER (\%)} & \textbf{ Kruskal-Wallis} & $I, D, S$ (\%) & \textbf{WER (\%)} & \textbf{ Kruskal-Wallis} & $I, D, S$ (\%)\\
     \hline
    \multirow{2}{*}{\textbf{Gender}} & Female & 13.6 & \multirow{2}{*}{\shortstack[l]{$H$ = 23.3, \\ $p$ \textless 0.001}} &$I=4.0,D=1.6,\mathbf{S=6.4}$ & 9.6& \multirow{2}{*} {\shortstack[l]{$H$ = 87.1, \\ $p$ \textless 0.001}} &$I=2.9,D=1.4,\mathbf{S=4.5}$ \\
\cline{2-3}\cline{5-6}\cline{8-8}
     & Male & \textbf{14.5} &  & $I=4.3,D=2.0,\mathbf{S=7.2}$ & \textbf{11.9} &  & $I=3.5,D=2.0,\mathbf{S=5.5}$ \\ 
     \hline\hline
     \multirow{2}{*}{\textbf{Caste}} & RES & 13.8 & \multirow{2}{*}{\shortstack[l]{$H$ = 15.2, \\ $p$ \textless 0.001}} & $\mathbf{I=4.1},D=2.0,S=7.0$ & 11.4 & \multirow{2}{*}{\shortstack[l]{$H$ = 16.4,\\ $p$ \textgreater 0.001}} & $\mathbf{I=3.4},D=1.9,S=5.4$ \\
     \cline{2-3}\cline{5-6}\cline{8-8}
     & UR & \textbf{14.6} & & $\mathbf{I=4.4},D=2.0,S=7.2$ &\textbf{11.9} &  & $\mathbf{I=3.5},D=1.9,S=5.5$\\ 
     \hline\hline
     \multirow{4}{*}{\textbf{Experience}} & $\leq$ 1980 & \textbf{15.9} & \multirow{4}{*}{\shortstack[l]{$H$ = 125.8,\\ $p$ \textless 0.001}} & $I=4.9,D=2.2,\mathbf{S=8.0}$ & \textbf{13.3} & \multirow{4}{*}{\shortstack[l]{$H$ = 135.8,\\ $p$ \textless 0.001}} & $I=3.9,D=2.1,\mathbf{S=6.3}$ \\
     \cline{2-3}\cline{5-6}\cline{8-8}
     & 1981-90 & 13.3 &  & $I=3.8,D=1.9,\mathbf{S=6.8}$& 11.5 &  & $I=3.3,D=1.8,S=5.4$\\ 
     \cline{2-3}\cline{5-6}\cline{8-8}
     & 1991-00 & 14.9 &  & $I=4.5,D=1.9,S=7.2$& 11.9 &  & $I=3.6,D=2.0,S=5.5$\\ 
     \cline{2-3}\cline{5-6}\cline{8-8}
     & $\geq$ 2001 & 14.0 &  & $I=4.2,D=1.9,S=6.9$ & 11.2 &  & $I=3.2,D=1.9,\mathbf{S=5.1}$\\ 
     \hline\hline
     \multirow{4}{*}{\textbf{Native}} & North & 13.6 & \multirow{4}{*}{\shortstack[l]{$H$ = 80.9,\\ $p$ \textless 0.001}} & $I=3.8,D=1.9,S=6.9$ & 10.8 & \multirow{4}{*}{\shortstack[l]{$H$ = 80.3,\\ $p$ \textless 0.001}} & $I=3.1,D=1.7,S=5.2$  \\
     \cline{2-3}\cline{5-6}\cline{8-8}
     & South & \textbf{15.4} &  & $\mathbf{I=4.6},D=2.1,S=7.5$ & \textbf{12.4} &  &$I=3.5,D=2.2,S=5.7$\\ 
     \cline{2-3}\cline{5-6}\cline{8-8}
     & East & 14.2 &  &$I=4.5,D=1.8,S=6.8$& 11.6 &  &$\mathbf{I=3.6},D=1.8,S=5.3$\\ 
     \cline{2-3}\cline{5-6}\cline{8-8}
     & West & 14.4 &  &$\mathbf{I=3.6},D=2.3,S=7.4$& 11.4 &  &$\mathbf{I=3.0},D=1.8,S=5.4$\\ 
     \hline\hline
     \multirow{3}{*}{\textbf{Speech rate}} & Slow & \textbf{15.3} & \multirow{3}{*}{\shortstack[l]{$H$ = 104,\\ $p$ \textless 0.001}} &$\mathbf{I=4.9},D=1.9,S=7.4$ & \textbf{9.3} & \multirow{3}{*}{\shortstack[l]{$H$ = 160.6,\\ $p$ \textless 0.001}} &$\mathbf{I=4.1},D=2.0,S=5.7$\\
     \cline{2-3}\cline{5-6}\cline{8-8}
     & Avg. & 14.2 &  &$I=4.5,D=1.9,S=6.9$ & 7.8 &  &$I=3.5,D=1.8,S=5.3$\\ 
     \cline{2-3}\cline{5-6}\cline{8-8}
     & Fast & 13.9 &  &$\mathbf{I=3.7},D=2.1,S=7.0$& 7.4 &  &$\mathbf{I=2.9},D=2.0,S=5.4$\\ 
     \hline\hline
    \multirow{2}{*}{\textbf{Discipline}} & Non-Eng. & 13.5 & \multirow{2}{*}{\shortstack[l]{$H$ = 110.8,\\ $p$ \textless 0.001}} & $\mathbf{I=3.7},D=1.9,S=6.9$ & 11.0 & \multirow{2}{*}{\shortstack[l]{$H$ = 128.9,\\ $p$ \textless 0.001}} &$\mathbf{I=3.0},D=1.9,S=5.1$\\
     \cline{2-3}\cline{5-6}\cline{8-8}
     & Engg. & \textbf{14.8} &  &$\mathbf{I=4.5},D=2.0,S=7.3$ & \textbf{12.2} &  &$\mathbf{I=3.7},D=2.0,S=5.6$\\ 
     \hline
    \end{tabular}
    \caption{WER corresponding to various attributes across categories and Kruskal-Wallis test results for transcripts generated by YouTube and Whisper ASRs. The highest word error rate in each category corresponding to each ASR is highlighted in bold. Except for caste category in Whisper ASR, significant difference in all sub-groups is observed as per the Kruskal-Wallis test results. $I,D,S$ (\%) column indicates median of insertion, deletion and substitution component of WER corresponding to each subgroup. Substitution error share in WER is highest for both YouTube and Whisper ASR. The disparity in subgroups in both ASRs is due to different WER components. The WER component with highest disparity corresponding to each category is highlighted in bold. E.g., in gender category for YouTube ASR the difference between median insertion, deletion and substitution error for male and female is 0.3, 0.4 and 0.8 respectively. This indicates that the substitution component is primarily responsible for the observed WER disparity between male and female speakers and the same is highlighted.}
    \label{tab:wer_comparison_tbl}
\end{table*}

\subsection{Disparity within attributes}
We now look at the YouTube and Whisper ASR performances for transcript generation across various demographic and non-demographic attributes.

\begin{figure*}[t]
    \centering
	~\begin{subfigure}{0.32\columnwidth}
		\includegraphics[width= \textwidth, height=3cm, keepaspectratio]{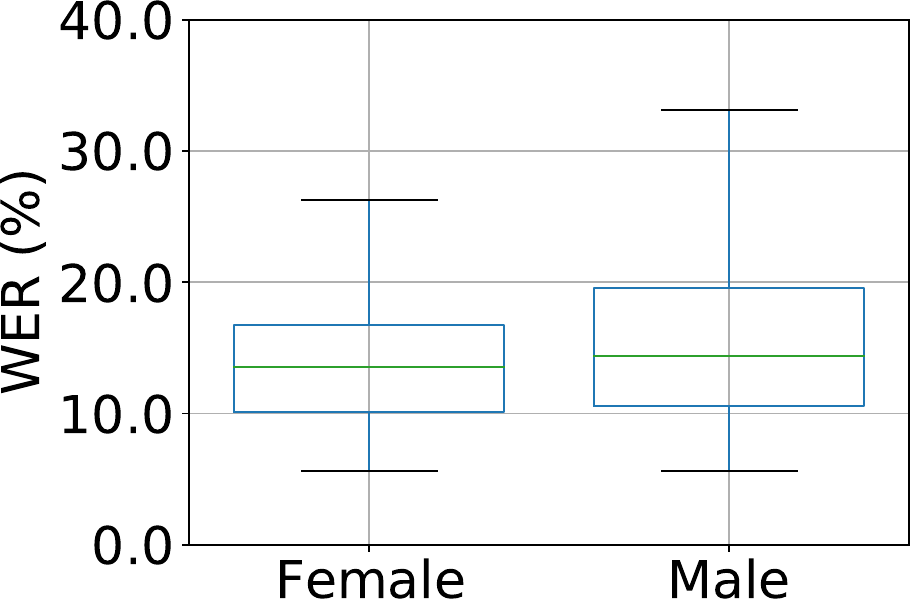}
		\caption{Gender}
		\label{fig:yt_gender}
	\end{subfigure}%
	~\begin{subfigure}{0.32\columnwidth}
		\includegraphics[width= \textwidth, height=3cm, keepaspectratio]{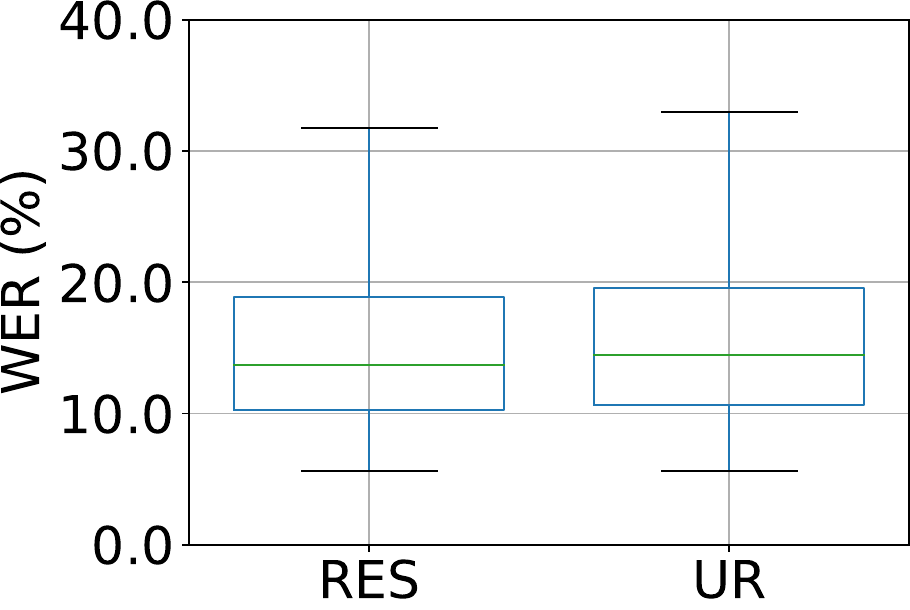}
		\caption{Caste}
		\label{fig:yt_caste}
	\end{subfigure}%
	~\begin{subfigure}{0.32\columnwidth}
		\includegraphics[width= \textwidth, height=3cm, keepaspectratio]{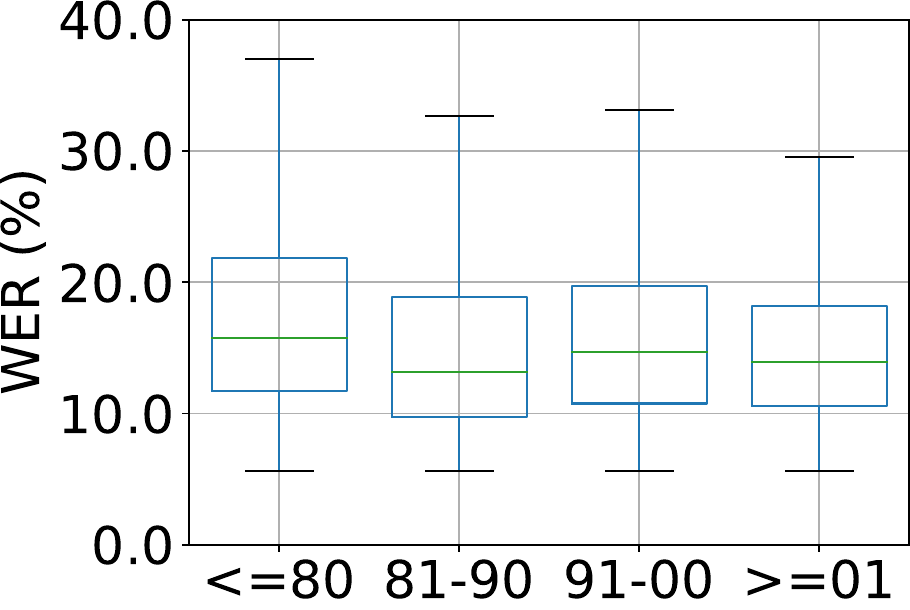}
		\caption{Experience}
		\label{fig:yt_exp}
	\end{subfigure} 
    \begin{subfigure}{0.32\columnwidth}
		\includegraphics[width= \textwidth, height=3cm, keepaspectratio]{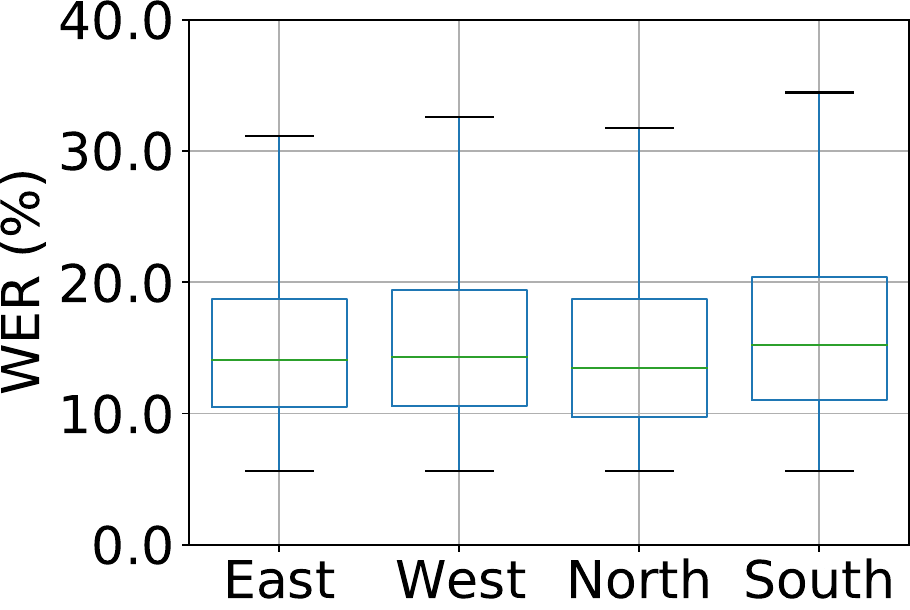}
		\caption{Native region}
		\label{fig:yt_native}
	\end{subfigure}
	~\begin{subfigure}{0.32\columnwidth}
		\includegraphics[width= \textwidth, height=3cm, keepaspectratio]{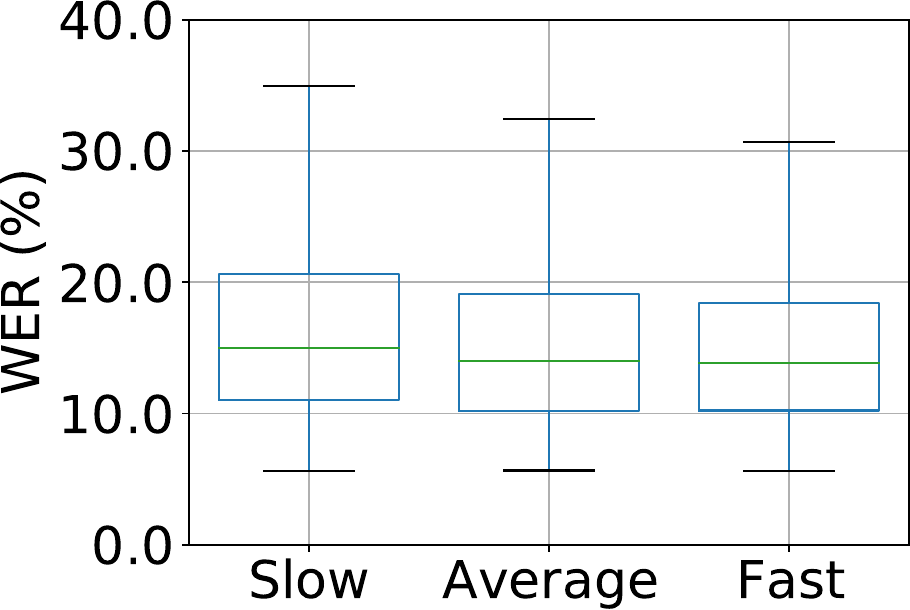}
		\caption{Speech rate}
		\label{fig:yt_speed}
	\end{subfigure}
	~\begin{subfigure}{0.32\columnwidth}
		\includegraphics[width= \textwidth, height=3cm, keepaspectratio]{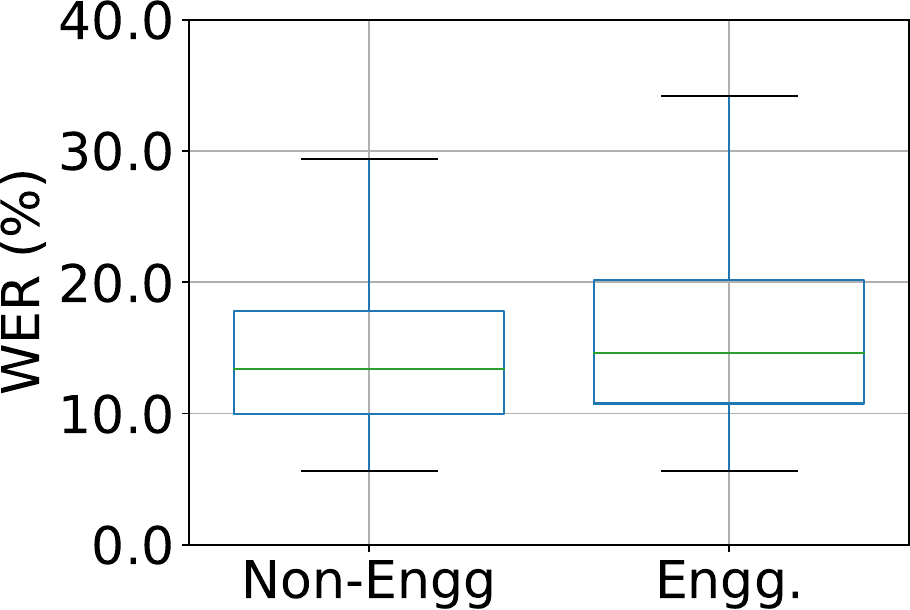}
		\caption{Discipline}
		\label{fig:yt_discipline}
	\end{subfigure}
 \hfill \textbf{YouTube}
 \\[2ex]
    ~\begin{subfigure}{0.32\columnwidth}
		\includegraphics[width= \textwidth, height=3cm, keepaspectratio]{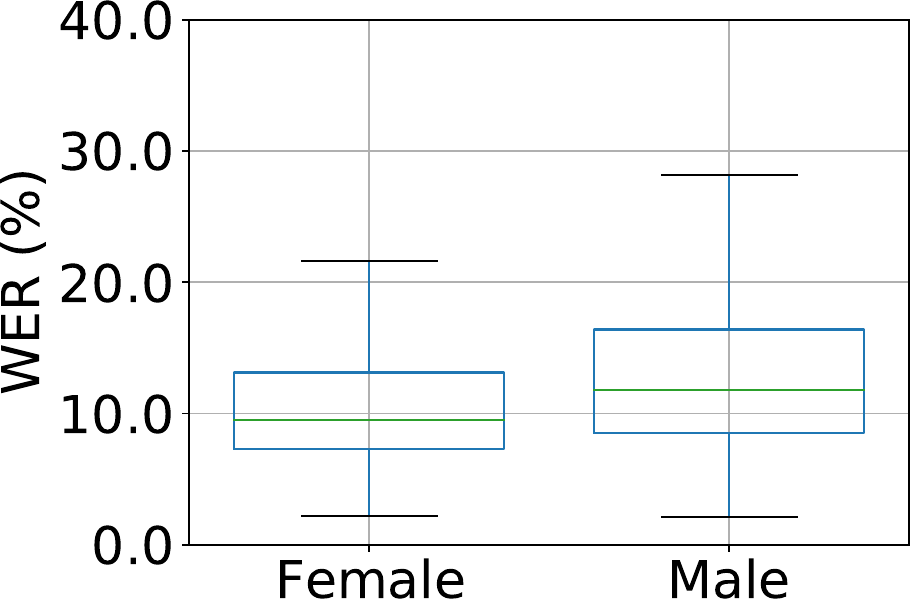}
		\caption{Gender}
		\label{fig:wsp_gender}
	\end{subfigure}%
	~\begin{subfigure}{0.32\columnwidth}
		\includegraphics[width= \textwidth, height=3cm, keepaspectratio]{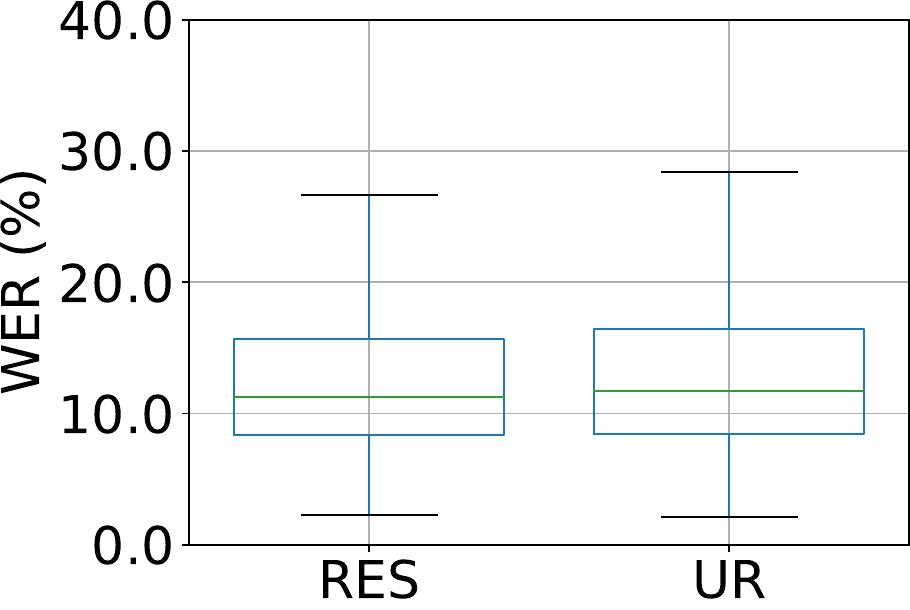}
		\caption{Caste}
		\label{fig:wsp_caste}
	\end{subfigure}%
	~\begin{subfigure}{0.32\columnwidth}
		\includegraphics[width= \textwidth, height=3cm, keepaspectratio]{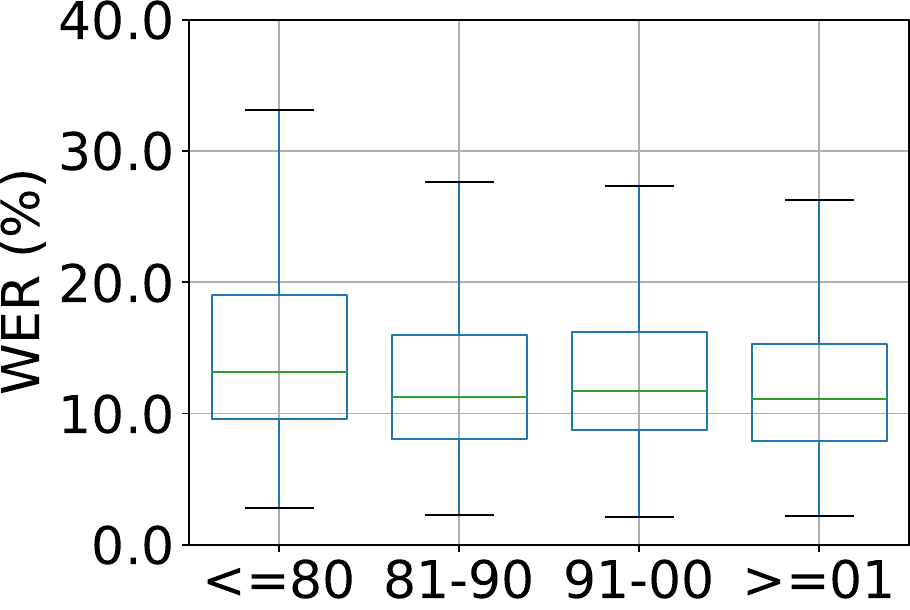}
		\caption{Experience}
		\label{fig:wsp_exp}
	\end{subfigure} 
    \begin{subfigure}{0.32\columnwidth}
		\includegraphics[width= \textwidth, height=3cm, keepaspectratio]{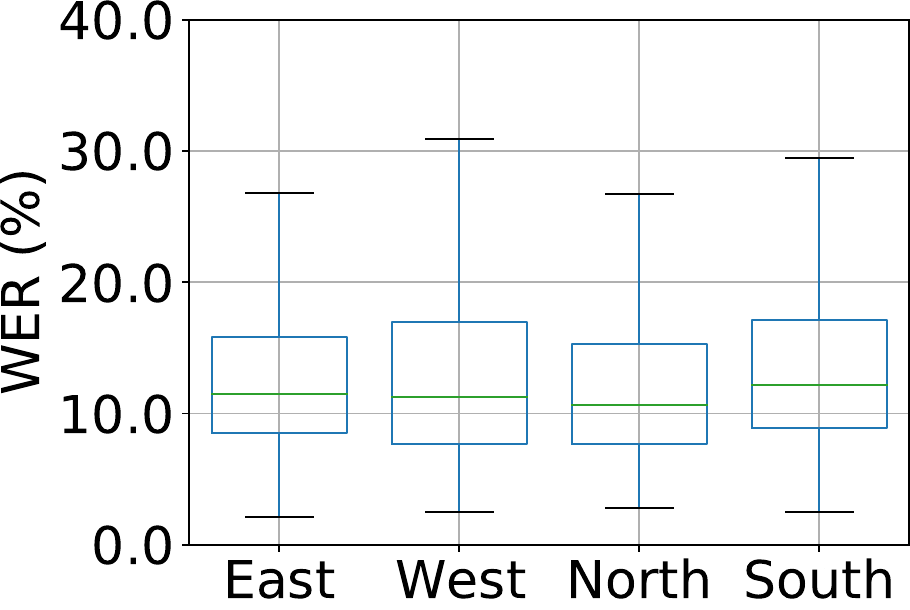}
		\caption{Native region}
		\label{fig:wsp_native}
	\end{subfigure}
	~\begin{subfigure}{0.32\columnwidth}
		\includegraphics[width= \textwidth, height=3cm, keepaspectratio]{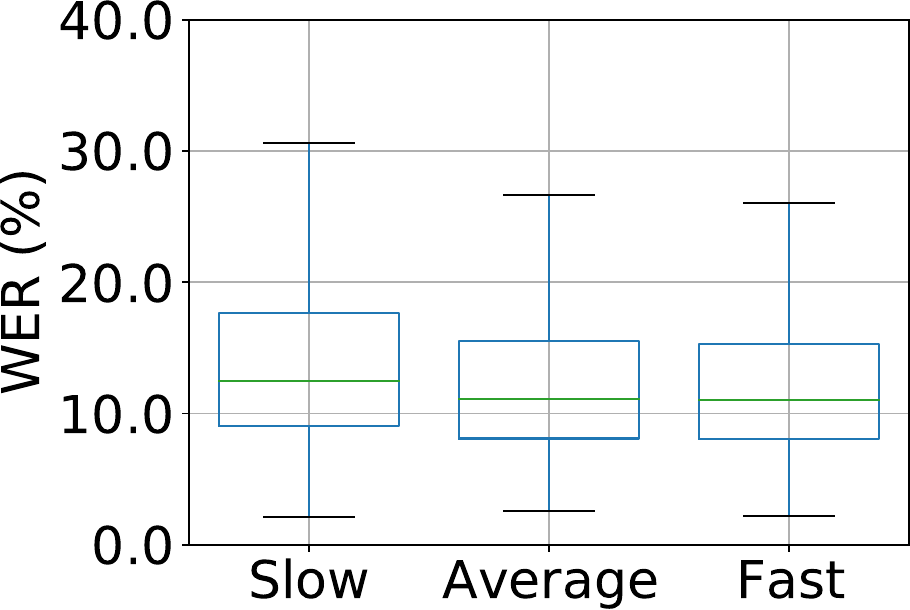}
		\caption{Speech rate}
		\label{fig:wsp_speed}
	\end{subfigure}
	~\begin{subfigure}{0.32\columnwidth}
		\includegraphics[width= \textwidth, height=3cm, keepaspectratio]{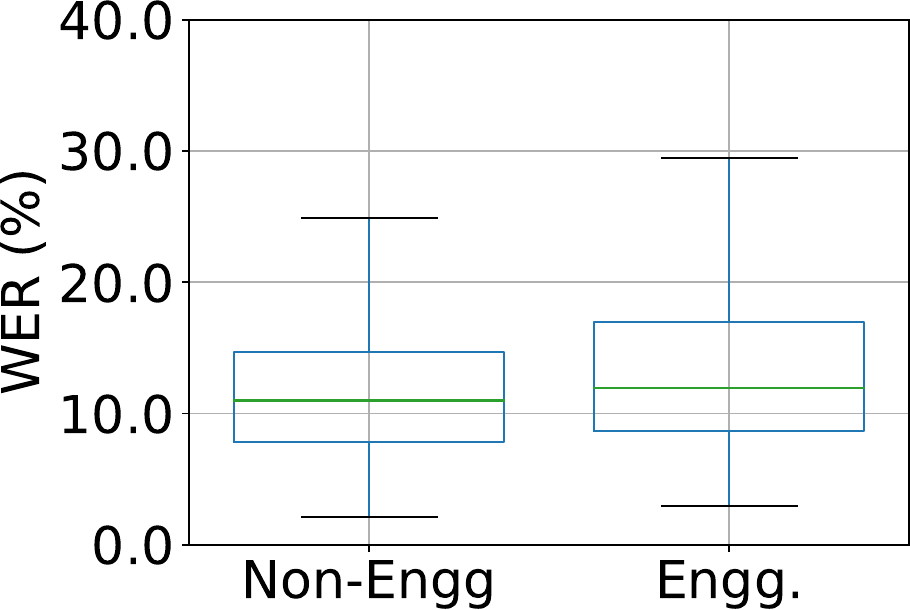}
		\caption{Discipline}
		\label{fig:wsp_discipline}
	\end{subfigure}
 \hfill \textbf{Whisper}
 \\[1ex]
	\caption{WER Disparity across various attributes for the YouTube and Whisper ASRs. The figures from (a) to (f) depict the distribution of WER through box plots for the YouTube ASR, while the figures from (g) to (l) show the distribution of WER through box plots for the Whisper ASR. }
	\label{fig:word_error_rate}
\end{figure*}

\subsubsection{Gender}
We compare the WER between male and female speakers in the TIE dataset. Even though the dataset has more than 95\% male speakers, we notice that female speakers report lower median WER from Table~\ref{tab:wer_comparison_tbl} (difference of 0.9\% for YouTube and 2.3\% for Whisper) and the  Kruskal-Wallis test's null hypothesis of no disparity in WER toward gender attribute can be rejected for both platforms as the $p$-value corresponding to test statistic is less than 0.001 (refer Table ~\ref{tab:wer_comparison_tbl}). Next, looking at the boxplot distributions in Figures~\ref{fig:yt_gender} and \ref{fig:wsp_gender}, we notice that males not only have a higher WER (34.0\% and 23.1\% resp.), but also a larger interquartile range (IQR) as compared to female speakers, irrespective of the platform. On comparing the two platforms, we see that Whisper reports lower median WERs (Table~\ref{tab:wer_comparison_tbl}) as well as IQR distributions and max WERs (Figs.~\ref{fig:yt_gender} and \ref{fig:wsp_gender}) for both genders, thereby performing better than YouTube for a given gender.

\noindent \textbf{Takeaways}: From the results for gender attribute, it can be inferred that --
\begin{itemize}
   \item WER for female voices is significantly lower (WER median difference -- 0.9\% on YouTube and 2.3\% on Whisper) than that of males, thus indicating a disparity between the transcripts for the two genders on both ASR platforms.
   \item Whisper performs better than YouTube for both genders, reporting a lower median WER and IQR, but it has a higher disparity between the male and female speakers as compared to YouTube.
   \item  The existing gender disparity in WER corresponding to both YouTube and Whisper is primarily due to substitution error disparity in the ground-truth and ASR-generated transcripts. 
\end{itemize}

\if{0}



\fi

\subsubsection{Caste}

On comparing the performance of speakers belonging to reserved castes against unreserved castes, we note for YouTube (see Table~\ref{tab:wer_comparison_tbl}), a significant median difference ($H$ = 15.2, $p\textless$ 0.001) of 0.8\%, with reserved caste speakers having low WER (median WER = 13.8\%) compared to unreserved caste speakers (median WER = 14.6\%). For Whisper on the other hand, we note that the difference is not statistically significant ($H$ = 16.4, $p\textgreater$ 0.001) and there is no disparity between the reserved and unreserved caste speakers. From Figures~\ref{fig:yt_caste} and \ref{fig:wsp_caste}, we see that the distributions are fairly similar but unreserved caste speakers report a slightly higher maximum WER at 33.9\% and 29.4\% respectively.

\noindent \textbf{Takeaways}:  On comparing the caste groups, it can be inferred that --
\begin{itemize}
    \item The Kruskal-Wallis tests shows that YouTube ASR has disparity between the reserved and unreserved caste category speakers, while Whisper does not. 
   \item Both platforms perform relatively better for reserved caste category speakers, reporting lower WER. 
   \item For both the ASRs, the WER dispersion in both groups is similar with difference in IQR for both groups being 0.8\%, indicating that the ASRs are agnostic to the caste of speakers.
   
\end{itemize}

\subsubsection{Experience}
Here we compare the ASR performance for speakers belonging to different experience groups.  From Table~\ref{tab:wer_comparison_tbl}, we notice that the worst performance is reported for the speakers with the highest experience (WER -- 16\% for YouTube and 13.3\% for Whisper). The best performance for YouTube is reported for speakers who started teaching in 1981-90 (WER = 13.3\%) and for Whisper, it was reported by the youngest speakers (WER = 11.2\%). Moreover, these differences are statistically significant for both the platforms. Interestingly, the worst WER for Whisper is the same as the best WER for YouTube thus showing the overall superiority of Whisper. 

Upon examining the video lectures of highly experienced teachers to uncover the reasons for high WER in their lectures, we discover that these teachers have a tendency to use filler words such as `uh', `um', `ok' and `alright' and repeat their last said words to bridge gaps between content words, which are part of the manual transcript. In addition, these experienced teachers often use blackboard teaching methods which can introduce noise into the speech signal due to the sound of writing on the blackboard and their movement while speaking. 

\noindent \textbf{Takeaways}: We now state the takeaways --
\begin{itemize}
   \item YouTube ASR performs best for speakers who started teaching between 1981-1990 and worst for those who started teaching before 1980. Table~\ref{tab:wer_comparison_tbl} highlights that speakers who began teaching before 1980 have the highest substitution error component for both YouTube and Whisper with values of 8.0\% and 6.3\% respectively. 
   \item Whisper ASR performs best for speakers who started teaching after 2001 and worst for those who started teaching before 1980, primarily due to substitution error.
   \item YouTube ASR favors the young speakers as the IQR reduces from 10.3\% to 7.8\% as we move from speakers with the most to the least experience (Figure~\ref{fig:yt_exp}). The IQR dispersion for Whisper ASR reduces from 9.8\% to 7.4\% for the same progression (Figure~\ref{fig:wsp_exp}).
   
\end{itemize}

\subsubsection{Native region}
We now study the change in the ASR performance based on the speaker's native region which may happen due to the regional variation in the accents. From Table~\ref{tab:wer_comparison_tbl}, we do not see any disparity between speakers hailing from west and east India in both ASRs, but both the ASRs perform best for speakers from north India, with median WER for YouTube being 13.6\% and median WER for Whisper being 10.8\%. Figure~\ref{fig:yt_native} shows the least IQR and maximum WER for speakers from east India, with the max values for south Indian speakers. 

\noindent \textbf{Takeaways}: The takeaways for the ASRs performance based on the speakers' native region are --
\begin{itemize}
    \item Whisper outperforms YouTube for speakers from all four regions.
    \item Both ASRs perform best for speakers from north India and worst for speakers from south India.
    
    \item  The primary cause of native regional disparity in WER is insertion error, which is highest between South and West Indian speakers in case of YouTube and East and West in case of Whisper. 
    
    \item The lowest IQR for 7.6\% and 8.5\% is reported for speakers from east India by Whisper and YouTube ASRs respectively.
\end{itemize}

\if{0}
The variation of median WER which is 14.2\%, 14.4\%,13.6\% and 15.4\% for native speakers from East, West , North and South region respectively, may be attributed to the variation in Indian English accent of speakers.
\fi

\subsubsection{Speech rate}
We now study the results for both ASRs for non-demographic attributes. In the TIE dataset, the share of slow, average and fast speech rate category lectures are as follows -- 39.1\%, 21.4\% and 39.5\% respectively. In Table~\ref{tab:wer_comparison_tbl}, we see that interestingly, the slowest speakers have the worst WER (15.3\% for YouTube and 9.3\% for Whisper) and the fastest speakers have the best WER (13.9\% for YouTube and 7.4\% for Whsiper). Figures~\ref{fig:yt_speed} and~\ref{fig:wsp_speed} show a similar change in IQR and it decreases with the increase in speaking speed.

\noindent \textbf{Takeaways}: We have the following takeaways for the two ASR performance for speakers with different speech rates --
\begin{itemize}
   \item Both ASRs are best in transcribing the audio files for speakers who speak the fastest.
   \item The performance for both ASRs become more consistent as we move from slow to fast speech rate (Figs.~\ref{fig:yt_speed} and \ref{fig:wsp_speed}).
   \item The highest disparity in WER between lectures with slow speech rate and lectures with fast speech rate can be attributed to the insertion error component of WER. 
\end{itemize}

\if{0}
There is a WER median difference of 1.4\% favoring datapoints belonging to Fast Category(median WER 13.9\%) compared Average(median WER 14.2\%) and Slow(median WER 15.3\%) Speech Rate Category. As the distribution of the dataset corresponding to speech rate categoy is evenly distributed making the median difference even more significant.
\fi

\subsubsection{Discipline}
We divide the lecture videos into two disciplines -- non-engineering and engineering to understand the differences in ASR performances between these two broad disciplines. From Table~\ref{tab:wer_comparison_tbl}, we observe that transcripts from engineering disciplines have a higher WER (difference of 1.3\% for YouTube and 1.2\% for Whisper) than those from non-engineering domain. This difference is statistically significant in both ASRs and thus a disparity exists between the two categories. From Figures~\ref{fig:yt_discipline} and~\ref{fig:wsp_discipline}, we see that YouTube's WER for engineering lecture videos have a higher IQR and maximum WER, thereby indicating an overall worse performance than Whisper. Table~\ref{discipline_tbl} illustrates the two disciplines with the best and worst WER for each ASR.


\begin{table}[!t]
    \centering
    \tiny
    \begin{tabular}{|c|c|c|c|c|}
        \hline
        \multirow{2}{*}{\textbf{ASR}} & \multicolumn{2}{c|}{\textbf{Lowest WER}} & \multicolumn{2}{c|}{\textbf{Highest WER}} \\
        \cline{2-5}
         & \textbf{Engg.} & \textbf{Non-Eng.} & \textbf{Engg.} & \textbf{Non-Eng.} \\
         \hline
         \multirow{2}{*}{YouTube} & Textile & Agriculture & Comp. Sci. & Chemistry \\
         \cline{2-5}
         & Mining & Basic Courses & Electronics Comm. & Mathematics\\
         \hline
         \multirow{2}{*}{Whisper} & Nano-Tech & Basic Courses & Comp. Sci. & Chemistry \\
         \cline{2-5}
         & Textile & Agriculture & Electronics Comm. & Atmos. Science\\
         \hline
    \end{tabular}
    \caption{Two disciplines having the highest and lowest WER for each of the disciplines and each of the ASRs.}
    \label{discipline_tbl}
\end{table}



\noindent \textbf{Takeaways}:  We now list the takeaways for comparison between the two disciplines --
\begin{itemize}
   \item Both ASRs perform better in lectures belonging to non-engineering category compared to the engineering category.
   \item In both YouTube ASR-generated transcripts and Whisper-ASR generated transcripts, the disparity in transcription accuracy between engineering and non-engineering discipline groups can be attributed to insertion errors.
   \item From Figs.~\ref{fig:yt_discipline} and~\ref{fig:wsp_discipline}, we see that engineering videos have a much higher maximum WER and a larger IQR for both ASRs, with YouTube performing worse than Whisper.
\end{itemize}


\if{0}

that there is disparity in its performance(H=110.8,$p\leq 0.001$) and it is performing better for non-engineering disciplines (median WER = 13.5\%) compared to engineering disciplines(median WER = 14.8\%).


\begin{table}[!t]
    \centering
    \small
    \begin{tabular}{|l|c|c|c|c|}
    \hline
    \multirow{2}{*}{\textbf{Speech Rate}} & \multicolumn{2}{c|}{\textbf{YouTube WER}} & \multicolumn{2}{c|}{\textbf{Whisper WER}}\\
    \cline{2-5}
    & \textbf{Engg.} & \textbf{Non-Eng.} & \textbf{Engg.} & \textbf{Non-Eng.} \\
    \hline
    Slow & \textbf{15.7} & 14.4 & \textbf{13.2} & 11.5 \\ 
    \hline
    Average & 14.8 & 12.6 & 11.9 & 10.0 \\ 
    \hline
    Fast & 14.2 & 13.0 & 11.2 & 11.0 \\ 
    \hline
    Kruskal-Wallis & \multicolumn{2}{c|}{$H$ = 223.2, $p\textless$ 0.001} & \multicolumn{2}{c|}{$H$ = 320.5, $p\textless$ 0.001} \\
    \hline
    \end{tabular}
\caption{WER comparison of YouTube and Whisper ASR for the intersection of speech rate and discipline.}
\label{tab:inter_speech_disc}
\end{table}

\begin{figure*}[!t]
	\centering
	\begin{subfigure}{0.75\columnwidth}
		\includegraphics[width= \textwidth, height=8cm, keepaspectratio]{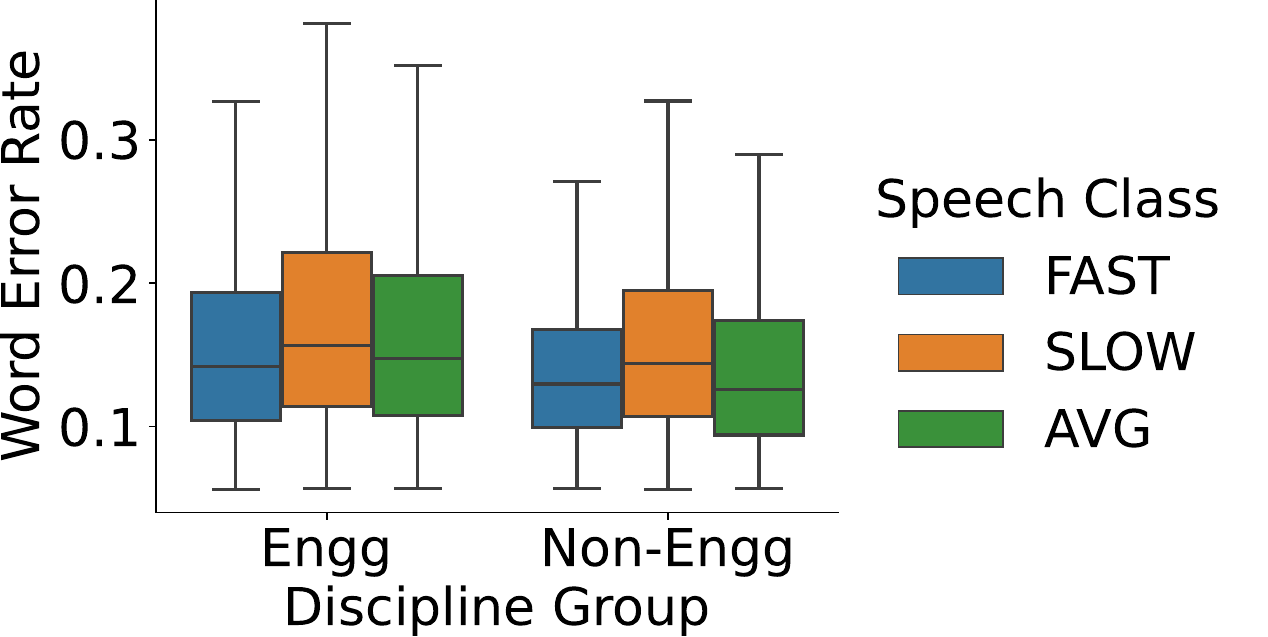}
		\caption{YouTube}
		\label{fig:yt_speech_discipline}
	\end{subfigure}
	~\begin{subfigure}{0.75\columnwidth}
		\includegraphics[width= \textwidth,  height=8cm, keepaspectratio]{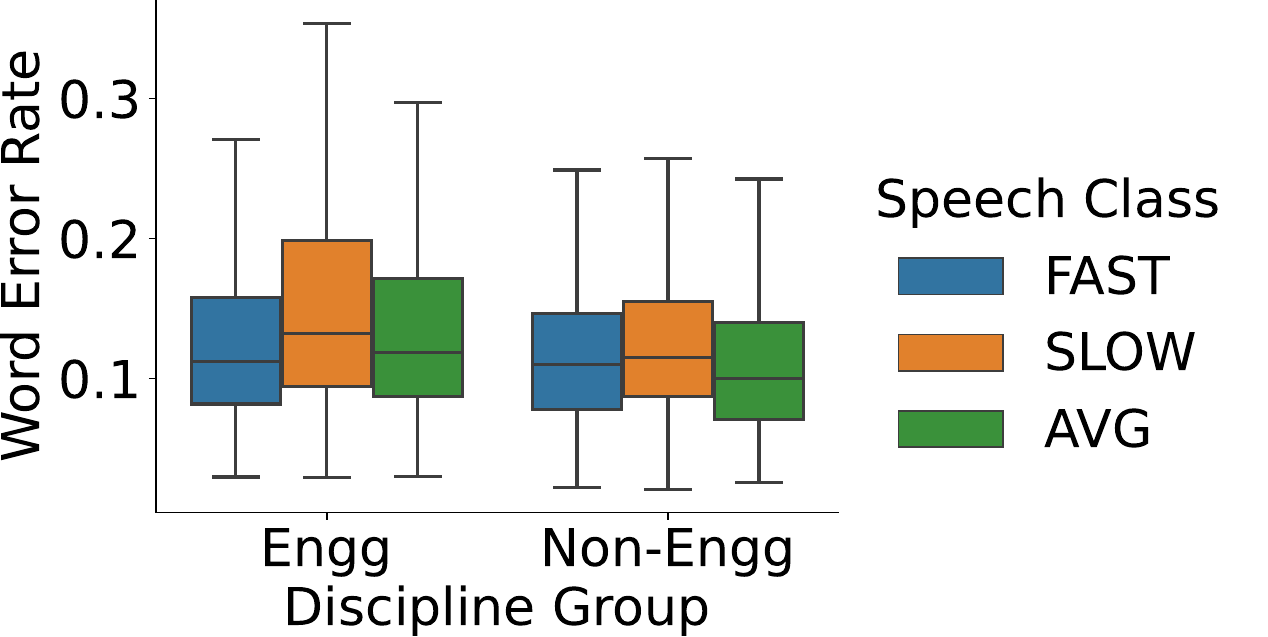}
		\caption{Whisper}
		\label{fig:wsp_speech_discipline}
	\end{subfigure}%
       \caption{WER disparity at the intersection of speech rate and discipline in YouTube and Whisper .\sjnote{Use patterns like -- 'x', 'o', '.' instead of colours. We haven't used colours elsewhere and we should be uniform.}}
       \label{fig:speech_discipline_comp}
 \end{figure*}

\subsubsection{Intersectional group}
We evaluate the performance of both ASRs for intersectional categories by combining speech rate and discipline. From Table~\ref{tab:inter_speech_disc}, we see that there is a statistically significant disparity in both ASRs for the intersectional categories. The difference in median WER between slow speech rate category lectures from engineering and average speech rate category of lectures from non-engineering discipline group is highest for both ASRs (3.1\% for YouTube ASR and 3.2\% for Whisper ASR).

\noindent \textbf{Takeaways}: We now state the takeaways from the intersectional experiments--
\begin{itemize}
   \item  Both ASRs have the worst WER for engineering lecture videos where the speakers have a slow speech rate.
   \item The least WER (12.6\% for YouTube ASR and 10.0\% for Whisper ASR) is reported for non-engineering lecture videos delivered with an average speech rate. 
   \item From Fig.~\ref{fig:yt_speech_discipline} and ~\ref{fig:wsp_speech_discipline}, we see that the box plots for the non-engineering category have consistently lower dispersion compared to their counterparts for the engineering category in both ASRs.
\end{itemize}
\fi
\section{Discussion}
In this section, we summarize our findings from our experiments on the two ASR platforms -- YouTube Automatic Captions and OpenAI Whisper. We evaluate these platforms on our large-scale dataset-- TIE, from the NPTEL MOOC website for word error rate and associated disparities across various demographic attributes of the speakers like gender, caste, experience, native region and non-demographic attributes like speech rate and discipline of the lecture content. We take a two-pronged approach to our analysis -- voice related attributes and content related attributes.

\subsection{Voice} 
We first look at the demographic attributes of the speakers like gender, caste, experience, native region (indicator of accent) and the speech rate that can be correlated to differences in voices of the speakers. 

\subsubsection{Gender}
The results for disparity toward gender for YouTube ASR in particular (a difference of 0.9\% in median WER between males and females), are surprising as a previous gender specific black-box audit of YouTube Automatic Captions \cite{tatman2017effects} did not report any significant differences. In fact, \cite{tatman2017gender} found that the YouTube ASR performs better for White male speakers on an English speakers dataset. Thus, we can see that the accent (non-native Indians vs native White speakers) may be impacting the transcription accuracy and deploying the same model without accounting for regional accents can have an adverse effect on the audience's experience.

The difference in pitch, intensity, tonality along with the enunciation of speech characterize the difference in male and female voices.~\cite{adda2005speech} have pointed out better pronunciation and articulation of word utterances in female speech helps the ASR model in extracting correct content information from the speech signal. This may explain the lower WER for female speakers on both YouTube and Whisper ASR. Whisper ASR improves the accuracy for female voices, resulting in a 4\% lower WER for female voices and 2.6\% lower WER for male voices when compared to YouTube ASR. However, it has been observed that Whisper ASR has a greater variation in performance based on the gender of the speaker.

We also observe that the highest disparity between the two gender groups for substitution error on both ASRs. This confirms that the models are picking up the correct phoneme from female utterances more often than that of male utterances.

These differences could be attributed to the model architecture as well as the training data.

\subsubsection{Caste}
Overall, the differences based on caste are marginal which is also expected. This reinforces the fact that one's caste does not play any role in their tonal attributes or the quality of delivery.

\subsubsection{Experience}
~\cite{vipperla2010ageing} have highlighted that the organs involved in speech production mechanism of individuals like lungs, vocal cords and the vocal cavities get affected with age which in turn affects the articulation of words.The speakers with the highest experience are also expected to be the oldest and vice versa. The differences in median WER of the least and most experienced speakers in both ASRs depicts the inability of the ASR model to account for these phonetic variations.Similarly, the highest disparity in substitution error of both ASRs points toward the same direction.

While Whisper reports the lowest WER for the least experience category speakers which is intuitive, interestingly, YouTube ASR reports the lowest value for speakers who joined their institutes between 1981-90.  Since the YouTube ASR is a black box model, it is difficult to reason about this anomaly.

 \subsubsection{Native region}  
Our results indicate a disparity between the word error rates for speakers belonging to different parts of India. ~\cite{pickering2000pitch} have studied the accents of Indian English discourse and have highlighted the difference in frequencies in accented/unaccented syllables of native Tamil (southern region), Bengali (eastern region) and Hindi/Urdu (northern region) speakers. The higher variation in accent of English spoken by the northern and southern speakers is reflected in the median WER difference for YouTube and Whisper which are 1.8\% and 1.6\% respectively. 

It should be noted that the high insertion error rate in both of the ASRs for South and East Indian speakers could be due to their accent having certain regional influences. The ASR system may be adding (sub)words that were not originally part of the spoken utterances.

\subsubsection{Speech rate}
The speech rate of native English speakers is understandably higher than that of non-native speakers~\cite{guion2000age}. This could be the most plausible explanation for better transcription for lectures having fast speech rate by both ASRs. Whisper ASR has been found to improve the accuracy of speech recognition for lectures delivered at slow, average and fast speech rates by 6\%, 5.4\% and 6.5\% respectively, when compared to YouTube ASR. Nevertheless, it has a greater variation in performance depending on the speech rate attribute in comparison to YouTube.

~\cite{wang2007robust} has highlighted the role of speech rate in sentence boundary, disfluency and syllable detection accuracy in speech. Longer syllable duration in lectures which are categorized in slow category might be affecting the correct syllable detection leading to the insertion of extra words in ASRs. The same has been observed in WER error components of both ASRs, where insertion error disparity is highest due change of speech rate. The insertion error rate is lowest for lectures delivered in fast speech rate while those delivered at a slower pace have a higher insertion error rate.  


To summarize, we observe statistically significant disparities in both ASRs due to voice related attributes of speakers except for caste where the differences are marginal.
 From these findings, it can be argued that the disparities observed in both Whisper and YouTube on the TIE dataset are not only due to imbalances in linguistic variations in the training set, but also due to the limitations in the model architecture that does not account for variations in speech signal features. As  OpenAI Whisper is an open source model, it can be fine-tuned for the speech variations based on the use case; however the same will not be applicable for YouTube.

\subsection{Content}
Next, we look at the content related attribute -- discipline to which the lecture videos belong. 

\subsubsection{Discipline}
Our findings on the disparity between the WER for engineering and non-engineering lectures points toward the lack of generalized performance on domain specific data sets.It is also observed that there exist disparities within each discipline as well and these may be an outcome of the difference in the number and type of subjects covered in each group. For example, in engineering discipline, average WER varies from 20\% for YouTube and 18\% for Whisper in Computer Science and Engineering to 12\% for YouTube and 10\% for Whisper in Nanotechnology. In non-engineering discipline, the average WER varies from 19\% for YouTube and 15\% for Whisper in Chemistry and Biochemistry to 10\% each in YouTube and Whisper in Agriculture.

Our findings suggest that Whisper outperforms YouTube in terms of reducing the WER for various voice and content related attributes. However, Whisper's zero shot transcription feature is observed to exhibit more disparities across certain attributes and groups, compared to YouTube. Despite this, open-source models like Whisper are still a better alternative in comparison to proprietary models in terms of performance, flexibility and addressing disparities. 

It is important to note that the model architecture should be designed to account for variations in speech signal features in order to further reduce disparities in the results. This could be achieved by having modern adversarial setups~\cite{siva20} and transfer learning  ~\cite{vu2014improving,shibano2021speech} that have been found to be effective in various problems.
The TIE dataset and other datasets ~\cite{sanabria2023edinburgh,ahamad2020accentdb,zhao2018l2} that include speech samples from non-native speakers of English can be valuable resources for improving the accuracy of open-source ASR models like Whisper. By fine-tuning these models on a diverse range of speech samples including those from speakers with accents, slower speech rates and other potential idiosyncratic attributes, it is possible to reduce the performance gap between different subgroups of speakers. Integrating audio preprocessing techniques such as noise filtering, speaker movement compensation and other similar methods into the ASR pipeline improves the signal quality and can also contribute to enhancing the overall performance of the ASR system. In addition, by incorporating an in-domain vocabulary set, the ASR model can be fine-tuned to recognize and transcribe speech in the target domain more accurately as shown by ~\cite{saadany2022better}.    
\section {Conclusion}

In this paper, we set out to explore the issue of disparity in ASR systems towards Indian demography. In order to accomplish the same we have proposed the TIE dataset comprising a set of technical lectures delivered by Indian speakers representing different dimensions of India's socially and linguistically diverse population. We benchmark the performance of YouTube and Whisper for measuring disparity due to speaker characteristics using the word error rate evaluation metric using this dataset. Additionally, we analyzed the insertion, substitution, and deletion components of WER to gain insights into the causes of errors in the evaluated ASR models.

Our findings highlight that there is a strong correlation between WER patterns for both ASRs being audited. Significant disparities across gender, speech rate, age group and native region of speakers exist for both the ASR systems. We also find that the disparity due to caste is marginal. Both the ASRs seem to have high error rates for lecture videos in the engineering disciplines compared to the non-engineering disciplines.


 We also discussed strategies like use of more diverse and representative training data for ASR model training and evaluation. It is important to build customized models that are specifically tailored to the speech variations of underrepresented groups and lexical variations of technical domains to mitigate the existing disparities. Adversarial and transfer learning methods could be very helpful in this context. 

\subsection{Potential limitations}

 Our study may have some limitations for example, the correctness of manual transcripts can be impacted by human error during transcribing. Similarly, self-annotated labels like gender and caste are not publicly available and hence challenging to determine. Automatic identification of these attributes from names might have some potential errors. The formal speech style in the videos and use of technical jargon may have an impact on the ASR performance. Some of the subjects involve live mathematical derivations and problem solving while others are delivered using slides. These factors may also affect the ASR performance favorably or adversely depending on the lecture topic. Also, the non-deterministic outputs of ASR model components may result is different transcripts of same video. These may all impact the performance. To address these, we experiment over a large dataset and consider a confidence interval of 99.999\% for statistical significance.  

\subsection{Broader perspective, ethics and competing interests}
ASR systems have been shown to have bias toward specific sub-population of society due to non-standard accent and dialects. For a linguistically diverse and vast population like India, bias of ASR systems may have serious consequences in providing equitable access to ASR based technologies. In this work, we evaluate disparities in two ASRs -- YouTube Automatic Captions and OpenAI Whisper -- on a self curated dataset of technical lectures, which is highly relevant from a pedagogical point of view in the Indian context. As online education has been an enabler in democratising education in developing economies like India, errors in ASR generated transcripts can seriously impact both lecture delivery of instructors and learning outcome of students~\cite{willems2019increasing}. 

This audit study and evaluation dataset will enable the ASR system developers in investigating and addressing the biases in these systems which disproportionately affect certain demographic groups. 
However, if our dataset is used for training any ASR system, it may exhibit data bias due to skewed representation of some of the demographic attributes. Thus we advise practitioners to use caution with the same.

While annotating sensitive demographic attributes of the speakers like gender and caste in our dataset, we specifically label them as perceived gender and perceived caste since self-identified gender and caste information were not available in the public domain. We take due care in anonymising the instructor name and associated institute name corresponding to each lecture to avoid individual level and institute level evaluation based on word error rates in future works.

\if{0}
\sjnote{Write this at the very end. We don't know what results will be added or not eventually. Right now this is too open-ended.}
We audited the transcription error bias in the acoustic and language model of YouTube Captions and OpenAI Whisper using 8740 hours of engineering lecture videos of various domains from 332 instructors representing different dimensions of Indian demographic structure.

From this study, it has been established that significant bias due to gender, caste, accent, speech rate, age group, native region of speakers does exist in Indian context also in both ASRs being audited. Further it is also found that there is a bias in word error rate in transcriptions of computer-assisted instructors and unassisted instructors. 

After analysing acoustic model and language model of both ASRs separately based on content and feature of speech, it is found that language model of both ASRs is robust to diverse content and overall performance differences between both ASRs arises primary due to their acoustic model.

Possible ways to reduce such biases are by re-training acoustic model of these systems with more acoustically diverse training samples having variation in speech style, speech rate and adjusting acoustic models to be better suited for speakers of different linguistic backgrounds and speech with varied auditory attributes.

\fi

\bibliography{references}
\end{document}